%% file: main.tex
\documentclass{article} 

\usepackage{iclr2026_conference,times}

\input{math_commands.tex}

\usepackage{hyperref}
\usepackage{url}
\usepackage{graphicx}
\usepackage{wrapfig}
\usepackage{subcaption}
\usepackage{booktabs}
\usepackage{multirow}
\usepackage{amsmath}
\usepackage{amssymb}
\usepackage{amsfonts}
\usepackage{mathtools}
\usepackage{amsthm}
\newtheorem{proposition}{Proposition}
\newcommand{\muq}{\bm{\mu}_Q}   

\usepackage{xcolor}
\usepackage{colortbl}
\usepackage{pifont}
\usepackage{nicefrac}
\usepackage{microtype}
\usepackage[capitalize,noabbrev]{cleveref}
\usepackage[font=small,labelfont=bf]{caption}

\usepackage[acronym,nomain]{glossaries}
\setacronymstyle{long-short}
\glsdisablehyper 
\newacronym{trm}{TRM}{Tiny Recursive Model}
\newacronym{ptrm}{PTRM}{Probabilistic Tiny Recursive Model}
\newacronym{hrm}{HRM}{Hierarchical Reasoning Model}
\newacronym{eqr}{EqR}{equilibrium recursion}
\newacronym{ptq}{PTQ}{post-training quantization}
\newacronym{qat}{QAT}{quantization-aware training}
\newacronym{lsq}{LSQ}{learned step-size quantization}
\newacronym{raq}{RAQ}{recursive-aware quantization}
\newacronym{ocp}{OCP}{Open Compute Project}

\providecommand{\std}[1]{{\scriptsize$\pm#1$}}
\newcommand{\yes}{\textcolor{teal!70!black}{\checkmark}}
\newcommand{\no}{\textcolor{red!60!black}{\ding{55}}}
\definecolor{ourshl}{rgb}{0.90,0.94,1.00}
\newcommand{\ourrow}{\rowcolor{ourshl}}
\newcommand{\lowacc}[1]{\textcolor{red!65!black}{#1}}
\newcommand{\zq}{z^{q}}      
\newcommand{\zfp}{z^{\mathrm{fp}}}
\newcommand{\yq}{y^{q}}
\newcommand{\yfp}{y^{\mathrm{fp}}}
\newcommand{\fblock}{f_\theta}   
\newcommand{\Qop}{\mathcal{Q}}   

\hypersetup{
    colorlinks=true,
    citecolor=blue!60!black,
    linkcolor=black,
    urlcolor=teal!60!black
}

\title{Quantizing Recursive Reasoning Models}

\author{%
Thorir Mar Ingolfsson$^{1}$ \And
Wajeeha Tahir$^{1}$ \And
Anna Tegon$^{1}$ \And
Lionnus Kesting$^{1}$ \AND
\makebox[\dimexpr\textwidth-2\tabcolsep][c]{Gamze İslamoğlu$^{1}$\qquad\qquad Luca Benini$^{1}$}\\[5pt]
\makebox[\dimexpr\textwidth-2\tabcolsep][c]{\normalfont $^{1}$Integrated Systems Laboratory, ETH Z{\"u}rich, Switzerland}\\[2pt]
\makebox[\dimexpr\textwidth-2\tabcolsep][c]{\normalfont Corresponding Author: \texttt{thoriri@iis.ee.ethz.ch}}
}
\iclrfinalcopy 

\begin{document}

\maketitle

\begin{abstract}
\input{sections/00_abstract.tex}
\end{abstract}

\input{sections/01_intro.tex}
\input{sections/02_background.tex}
\input{sections/03_timestep_accumulation.tex}
\input{sections/04_blockwise_integer.tex}
\input{sections/05_recursion_structure.tex}
\input{sections/07_discussion.tex}

\subsubsection*{Reproducibility statement}
The completed experiments are evaluated on the public Sudoku-Extreme and Maze-Hard benchmarks using models trained with the released TRM codebase and the quantization code described in \Cref{sec:blockwise}, and on ARC-AGI-1/2. Exact configurations, seeds, and per-step logging scripts are provided in the supplementary material.

\bibliography{references}
\bibliographystyle{iclr2026_conference}

\appendix
\input{sections/99_appendix.tex}

\end{document}

%% file: math_commands.tex
\usepackage{amsmath,amsfonts,bm}

\def\eqref#1{equation~\ref{#1}}

\def\1{\bm{1}}

\DeclareMathAlphabet{\mathsfit}{\encodingdefault}{\sfdefault}{m}{sl}
\SetMathAlphabet{\mathsfit}{bold}{\encodingdefault}{\sfdefault}{bx}{n}

\newcommand{\E}{\mathbb{E}}



%% file: sections/00_abstract.tex
Recursive reasoning models solve hard puzzles by applying compact, weight-tied blocks over many refinement steps. Because these blocks are reused many times, quantizing them creates a unique dynamical problem: the quantization error is incurred at every step. While 8-bit quantization (integer or float) preserves accuracy, moving to a per-tensor 4-bit format causes a systematic bias to accumulate. The ensuing drift catastrophically degrades exact-solution accuracy on Sudoku from $84.1\%$ to $0.0\%$ (only ${\sim}25\%$ of cells correct). In this work, we show that this collapse is caused by activation-scaling \emph{granularity} rather than bit-width or number format. Crucially, moving to per-block scaling completely restores the transition. To implement this, we apply MXInt4, a blockwise integer activation format, to recursive reasoning models. It is competitive with blockwise float formats on our tasks, while keeping integer elements and power-of-two block scales. Finally, recursion depth and reuse modulate quantization sensitivity, with the deepest architecture we test (the EqR equilibrium model) the most sensitive. Yet blockwise scaling overcomes this vulnerability, staying robust across these architectures and transferring to the open-ended ARC-AGI benchmark.

%% file: sections/01_intro.tex
\section{Introduction}
\label{sec:intro}

A recent line of research has shown that hard reasoning problems can be solved by organizing computation recursively with few parameters. The \gls{hrm} \citep{wang2025hrm} and the \gls{trm} \citep{jolicoeur2025trm} reach competitive accuracy on Sudoku, maze solving, and ARC-AGI by applying compact, weight-tied transition functions over many refinement steps. This line of work has already expanded to stochastic, autoregressive, hybrid, long-horizon, and language-model variants \citep{parviz2026ptrm,baek2026gram,tarm2026,mambatrm2026,yang2026longhorizon,sapient2026hrmtext}. These models are attractive for low-precision inference precisely because their parameter count is small. Yet their behavior under quantization remains essentially unstudied: they are reported almost entirely in full precision, without an accompanying low-precision accuracy or trajectory analysis.

To understand quantization behavior, we view a recursive reasoner as a learned state-space computation. Given an input $x$, a shared transition operator repeatedly refines a latent reasoning state and an answer state. Quantization fundamentally replaces this operator with a perturbed or noisy transition. 

In a standard feedforward model, a quantizer perturbs a finite sequence of \emph{different} layers. In a weight-tied recursion, however, the \emph{same} quantized transition is reused many times. This makes the induced trajectory just as important as the one-step error. While the recursion might safely absorb an approximately zero-mean perturbation, a systematic bias behaves differently. It can push the trajectory, or the fixed point it converges to, in a coherent direction at every step, resulting in a strong accumulated drift.

This paper studies quantization at the level of these reused transition operators. We examine when a low-bit transition stays faithful under repeated application, and how to build one that does. We make three primary contributions:

\paragraph{1. Quantization as transition perturbation.} We compare a full-precision transition with its quantized counterpart applied to the same reasoning state (\Cref{sec:timestep}). Empirically, 8-bit integer precision acts as a small perturbation. The quantized trajectory contracts back toward the full-precision path and accuracy is preserved. Conversely, per-tensor 4-bit quantization induces severe trajectory drift, dropping Sudoku exact accuracy from $84.1\%$ to $0.0\%$. The displacement concentrates in the most-reused recursive layers ($336\times$ per forward pass) and scales directly with reuse depth.

\paragraph{2. A blockwise integer transition rule.} We show that activation scaling \emph{granularity} is the dominant control variable (\Cref{sec:blockwise}). Weight-only 4-bit quantization costs only a few points. However, when quantizing activations at 4-bit, both per-tensor integer (Int4, $0.0\%$) and float (FP4, $0.6\%$) formats lose exact accuracy. In contrast, per-block scaling completely preserves the transition. We apply this principle using MXInt4, a blockwise activation format with integer mantissas, which reaches $80.1\%$ (Sudoku) and $84.7\%$ (Maze), competitive with blockwise float formats on these tasks while keeping integer elements and power-of-two block scales. Furthermore, tested outlier controls (SmoothQuant, Hadamard rotation) fail to provide this same post-training stability.

\paragraph{3. Recursion depth and reuse modulate quantizability.} We demonstrate this across an axis of architectures (\Cref{sec:structure}). The deepest, \gls{eqr} is the most sensitive. Per-tensor Int4 prevents attractor convergence, and stochastic breadth cannot average the systematic error away. Yet, our per-block rule restores stability and preserves depth scaling ($0.01\%\!\to\!82.3\%$). At the stochastic end of the axis, injecting test-time noise does not improve Int4 performance. This confirms the error is systematic rather than averageable. Finally, this same per-block rule successfully transfers to the open-ended ARC-AGI benchmark. We also show that \gls{qat} offers a complementary weight-side route towards even better accuracy (\Cref{sec:blockwise:arc,sec:blockwise:lsq}).

\begin{figure}[t]
  \centering
  \includegraphics[width=0.80\linewidth]{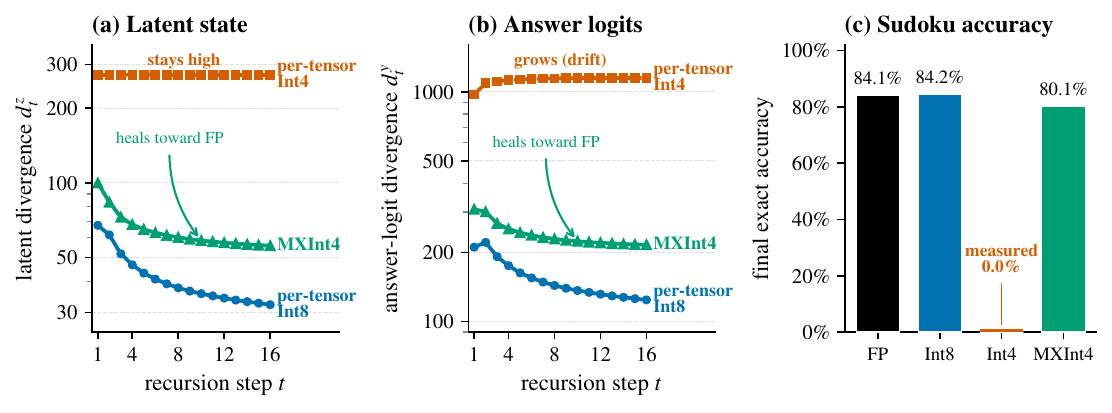}
\caption{\textbf{Transition contraction versus drift (Sudoku).} Quantized-vs-full-precision divergence over the $16$ recursion steps. \textbf{(a)} latent and \textbf{(b)} answer-logit divergence (log $y$-axis): per-tensor Int4 stays high in latent space and grows in logit space (a coherent drift), whereas 8-bit \emph{and} per-block MXInt4 both \emph{contract} back toward the full-precision trajectory. \textbf{(c)} final exact accuracy: per-tensor 4-bit removes every exact solution (measured $0.0\%$), while Int8 ($\approx$FP) and per-block MXInt4 ($80.1\%$; the fix developed in \Cref{sec:blockwise}) recover near-FP accuracy. Int8 is also per-tensor, so the failure is specific to per-tensor \emph{4-bit}, which per-block MXInt4 then repairs. The drift is milder on the shallower Maze recursion (${\sim}3$ steps; per-tensor Int4 $77.7\%$ vs.\ FP $84.7\%$ at this coverage).}
  \label{fig:hero}
\end{figure}

%% file: sections/02_background.tex
\section{Background and related work}
\label{sec:related}

\paragraph{Recursive and looped reasoning models.}\label{sec:related:rrm}

A recent line of work solves hard puzzle and reasoning tasks using tiny, recursively applied networks. For example, the \gls{hrm} \citep{wang2025hrm} couples two weight-tied recurrent modules running at different frequencies. The \gls{trm} \citep{jolicoeur2025trm} simplifies this to a single two-layer, ${\sim}7$M-parameter block. By recursing this block many times, it reaches $45\%$ on ARC-AGI-1 using a fraction of a percent of an LLM's parameters. These architectures share a lineage with weight-tied recurrent transformers \citep{dehghani2019universal} and looped transformers \citep{giannou2023looped,Xu2024loopedexpressive}. They control the number of shared-block applications using adaptive-compute mechanisms \citep{graves2016act}. Conceptually, they also parallel deep equilibrium models \citep{bai2019deq}, which iterate a weight-tied block to a fixed point.

This architectural space is growing quickly. Stochastic variants use test-time noise or learned per-step transitions to explore multiple solution trajectories \citep{parviz2026ptrm,baek2026gram}. Autoregressive variants morph standard transformers into shared-block recursion \citep{tarm2026}. Adjacent works incorporate curriculum learning, state-space hybrids, and long-horizon problem decomposition \citep{cgar2025,mambatrm2026,yang2026longhorizon}. Recently, HRM-Text \citep{sapient2026hrmtext} even applied this recursive structure to efficient language-model pretraining. Despite this rapid expansion, the literature on these models consists exclusively of full-precision studies. Our work addresses the missing interaction between quantization and recursion structure.

\paragraph{Quantizing weight-tied and recurrent networks.}\label{sec:related:rnnquant}

The difficulty of quantizing reused weights is a well-known effect: recurrent quantization is markedly more brittle than feedforward, with RNN binarization failing where low-precision feedforward networks succeed \citep{ott2016rnnquant,hubara2018qnn}, motivating RNN-specific quantizers and 4-bit LSTM pipelines \citep{he2016effectivernnquant,alom2018rnnquant,fasoli2021lstm4bit}. A recurring theme is that error grows with the number of passes through the reused weights. Precision Highway manages this accumulation \citep{park2018precisionhighway} and accumulator-aware quantization bounds the width needed to sum quantized terms safely \citep{colbert2023a2qoverflow,colbert2023a2q}. Theory on cross-layer weight sharing \citep{lan2020albert} and infinite-depth equilibria \citep{bai2019deq,moneq2026} grounds our claim, but the accumulation effect has never been measured on modern recursive reasoners. We focus on this characterization: the contraction-versus-drift behavior of quantized trajectories and its dose-response to reuse depth.

\paragraph{Iterated models: diffusion quantization as the closest cousin.}\label{sec:related:accumulation}

The closest analogue to our setting is quantized diffusion, where per-step error compounds across an iterated shared-weight denoiser and \gls{ptq} perturbs an iterated transition kernel rather than independent layers \citep{shang2023ptq4dm,li2023qdiffusion,he2023ptqd}. Our setting differs in two ways. First, a diffusion step \emph{conditions on the timestep} $t$, enabling $t$-dependent quantization parameters \citep{so2023tdq,huang2024tfmqdm}, whereas a weight-tied reasoner re-applies one operator with step-independent parameters. Second, diffusion drifts on a \emph{perceptual} manifold that tolerates small deviations, while a reasoner's solution manifold is \emph{all-or-nothing}: a single displaced symbol zeros out exact accuracy. Our per-tensor Int4 result is the logical-space analogue of this manifold drift. We expand the comparison, and the related exposure-bias literature \citep{ning2023exposurebias,arora2022exposurebias,he2021quantifyingexposure}, in \Cref{app:diffusion}.

\paragraph{Quantizing reasoning models, and low-bitwidth number formats.}\label{sec:related:reasoning}

Reasoning is highly sensitive to quantization. Empirical studies and broad \gls{ptq} benchmarks show that 8-bit reasoning LLMs are largely lossless. However, lower bit-widths become increasingly risky as task difficulty rises \citep{liu2025quanthurtsreasoning,zhao2025ptqbenchmark}. Fine-grained analyses of mathematical reasoning have localized these quantization-induced failures. Process-level attribution demonstrates that a single early faulty step can cascade and ruin an entire chain of thought \citep{li2025quantizationmeetsreasoningexploring,li2026quantreasoningcascade}. Recent work diagnoses and mitigates this using \gls{qat}, lookahead losses, and targeted protection of critical weights \citep{lv2026lowbitqat,quantlrm2026,laquant2026,zhang2026reasoningmeetscompressionunderstanding}. Weight-tied recursion represents the sharpest instance of this fragility. The exact same reasoning block must remain faithful across thousands of reuses.

On the format side, microscaling (MX) assigns a shared exponent to small blocks \citep{rouhani2023mxformats,ocp2023mxspec}. The LLM consensus is that 4-bit MXFP is lossy and an open challenge, with narrow dynamic range degrading accuracy \citep{huawei2026mxfpbenchmark,isfinerbetter2026}. Closing this gap needs machinery such as overflow-aware scaling, rotations, or metadata-augmented formats \citep{unveilingmxfp4_2026,bridginggapmxfp4_2026,lee2025amxfp4,hu2026m2xfp,blockrotation2025,batquant2026,arcquant2026}, and training recipes can make MXFP4/NVFP4 near-lossless via stochastic rounding or oscillation control \citep{tseng2025trainmxfp4,castro2025quartet,chmiel2025fp4alltheway,nvidia2025nvfp4,chen2025tetrajet}. Most relevant is that the integer-vs-float crossover is regime-dependent: MXInt often exceeds MXFP at 8-bit, while float usually leads at 4-bit unless paired with the right rotations \citep{chen2025intvsfp,zhang2023mofq,ashkboos2024quarot,liu2024spinquant}. This motivates our comparison. We find per-block \emph{integer} is competitive with per-block float on tiny weight-tied reasoners. This is appealing given the deployment literature's preference for integer-friendly edge formats and block-scaled accumulation \citep{quanttrim2025,zhuo2022tinymlquant,zhang2022fast}. Our building blocks are standard tools: learned quantizer parameters \citep{esser2020lsq,bhalgat2020lsqplus,choi2018pact}, the straight-through estimator \citep{bengio2013ste}, and \gls{ptq}/\gls{qat} baselines \citep{frantar2023gptq,xiao2023smoothquant,nagel2020adaround,li2021brecq,nagel2021whitepaper}. The closest combined work, BitSkip \citep{bitskip2025}, composes quantization with early exit layer-by-layer, whereas we frame accumulation across the \emph{recursion steps} of one shared block.

%% file: sections/03_timestep_accumulation.tex
\section{Quantization bias in reused transition operators}
\label{sec:timestep}

\subsection{Recursive transitions and quantized trajectories}
\label{sec:timestep:setup}

A weight-tied recursive reasoner repeatedly applies a single learned operator to refine a latent state. Quantization replaces this operator with a perturbed version at every reuse. This raises a key question: does the quantized trajectory stay close to the full-precision path, or does a small one-step perturbation compound into a coherent trajectory-level drift? We study the \gls{trm} \citep{jolicoeur2025trm}, which refines a latent state $z$ and answer embedding $y$ by applying a single two-layer, ${\sim}7$M-parameter block $\fblock$. This block is applied over $T$ deep-supervision steps consisting of several inner cycles. In our Sudoku configuration the inner ``$L$-level'' layers are applied $336$ times per forward pass. In contrast, the readout heads are applied only $16$ times. We evaluate on \emph{Sudoku-Extreme} (hard $9\times9$ puzzles as length-$81$ token sequences) and \emph{Maze-Hard} ($30\times30$ start-to-goal path-finding) \citep{wang2025hrm,jolicoeur2025trm}, scored by \emph{exact accuracy}: an instance counts only if every cell or path-step is correct (\Cref{app:config}).

Writing the recursive state $s_t=(z_t,y_t)$ and the full-precision recursion $s_{t+1}^{\mathrm{fp}}=F_\theta(s_t^{\mathrm{fp}},x)$, the quantized recursion is
\begin{equation}
  s_{t+1}^{q} = F_\theta(s_t^{q}, x) + \epsilon_Q(s_t^{q}, x).
  \label{eq:q-transition}
\end{equation}
Here, $\epsilon_Q$ is the induced input--output error of the \emph{entire} quantized transition, including all internal weight and activation quantizers. We do not assume this error is simply additive output-level noise. What matters is whether its mean component aligns across steps. At each step we therefore track task accuracy alongside two trajectory divergences:
\begin{equation}
  d^z_t = \lVert \zq_t - \zfp_t \rVert, \qquad d^y_t = \lVert \yq_t - \yfp_t \rVert.
  \label{eq:divergence}
\end{equation}
The shape of these divergence curves distinguishes a stable perturbation (where the divergence contracts) from a biased one (where displacement is reintroduced at every step). Unless noted otherwise, reported numbers are final-step point estimates. The per-tensor \gls{ptq} rows (\Cref{tab:transition-quantizers}) and the \gls{qat}/\gls{raq} comparison (\Cref{tab:lsq}) report mean$\pm$std over three seeds. Per-block (MX) rows are deterministic given the checkpoint. Per-step trajectory diagnostics quantize only the recursive block's linear layers, so their endpoints differ slightly from the broader-coverage ladder rows of \Cref{tab:transition-quantizers} (\Cref{app:config}).

\subsection{Low-bias perturbations contract; biased 4-bit perturbations drift}
\label{sec:timestep:signflip}

\Cref{fig:hero} presents our central trajectory observation. At 8-bit precision (Int8), the quantized transition behaves as a low-bias perturbation. The divergence is largest at the first step but \emph{shrinks} monotonically as the recursion proceeds. Consequently, final accuracy matches full precision (Sudoku Int8 $=84.2\%$ vs.\ FP $=84.1\%$ in \Cref{fig:hero}; Maze Int8 $=83.6\%$ vs.\ FP $=84.7\%$ in \Cref{tab:transition-quantizers}). Each application of the transition contracts the perturbed state back toward the full-precision trajectory faster than the quantizer displaces it.

At per-tensor 4-bit the sign of this effect completely flips. The trajectory divergence now \emph{grows} with depth. On Sudoku, the answer-logit divergence climbs from ${\sim}970$ at the first step to ${\sim}1150$ by step~16. The latent divergence saturates at a large value ${\sim}270$ and fails to decay. As a result, per-tensor Int4 \gls{ptq} drives exact accuracy from $84.1\%$ down to $0.0\%$. The same operator that damped 8-bit error now compounds 4-bit error: the quantized transition ceases to be a stable approximation and instead induces a coherent drift. On the easier Maze task (which converges in roughly three steps), the effect is milder. Int4 reaches $77.7\%$ versus $84.7\%$ at full precision, a ${\sim}7$-point drop. The fundamental asymmetry remains the same: 8-bit divergence decays while per-tensor 4-bit does not. Previewing \Cref{sec:blockwise}, the per-block 4-bit fix (MXInt4 in \Cref{fig:hero}) contracts like 8-bit, so scaling \emph{granularity} rather than bit-width sets the regime.

This behavior aligns with earlier evidence that recurrent quantization is more brittle than feedforward quantization \citep{ott2016rnnquant,hubara2018qnn} and that per-step error can accumulate across iterated computation \citep{park2018precisionhighway,arora2022exposurebias}. What is new is the sharp, controllable transition between contraction and drift on a model that reuses identical weights and quantization parameters at every step.

\subsection{Transition bias versus zero-mean noise}
\label{sec:timestep:mechanism}

Viewing quantization as a transition perturbation suggests a simple failure mechanism. A quantizer whose grid is poorly matched to the activation distribution produces an expected rounding error $\mathbb{E}[\Qop(a)-a]\neq 0$. In a feedforward network these biases occur across different, independent layers and need not align. In a weight-tied recursion the same transition is reused, so the mean component of $\epsilon_Q$ is injected in a similar direction at every application and the resulting displacement scales with the number of reuses. This behaves like a biased perturbation, which is fundamentally different from zero-mean noise.

We confirm the localization directly by calibrating an Int4 model and measuring the relative activation quantization error layer-by-layer (\Cref{fig:mechanism}). The error \emph{magnitude} is dominated by the recursive down-projection layers (relative error ${\sim}1.1$--$1.65$), exactly the layers re-applied hundreds of times per forward pass: the largest transition perturbations sit precisely where the recursion reuses them most. While \Cref{fig:mechanism} shows magnitude, the perturbation's \emph{systematic} nature is established behaviorally in \Cref{sec:structure:stochastic} and measured directly in \Cref{sec:blockwise:da1}: averaging independent trajectories and adding stochastic breadth fail to lift Int4 accuracy above zero, the signature of a coherent bias (consistent with the nonzero signed per-layer means found during calibration).

This distinguishes our setting from its closest cousin, diffusion-model quantization. A diffusion denoising step \emph{conditions on the timestep $t$}. This means accumulation can be damped with $t$-aware quantizers and calibration \citep{he2023ptqd,so2023tdq,huang2024tfmqdm}. A weight-tied \gls{trm} has no analogous $t$: it re-applies one operator with a single set of quantization parameters, which makes the accumulation more direct to analyze but larger in magnitude (extended comparison in \Cref{sec:related:diffusion}).

\subsection{A dynamical-systems perspective}
\label{sec:timestep:theory}

The contraction-versus-drift dichotomy admits a lightweight, conditional model (Proposition~1, a qualitative displacement scaffold rather than a tight bound, with the full statement in \Cref{app:theory}). We separate the per-step perturbation $\epsilon_Q$ of \Cref{eq:q-transition} into its magnitude (bounded by $\varepsilon$), its state variation (a \emph{smoothed} local sensitivity $\delta$, since piecewise-constant quantizers are not literally Lipschitz), and its trajectory mean $\muq(x)$. Motivated by the contracting Int8 divergence (\Cref{fig:hero}), we model the full-precision transition as locally contractive with modulus $L<1$. Two conclusions follow:
\begin{itemize}
    \item \textbf{Displacement:} if $L+\delta<1$ the perturbed transition still contracts, but to a fixed point displaced by at most $\varepsilon/(1-L)$, small at 8-bit and catastrophic at per-tensor 4-bit.
    \item \textbf{Coherent accumulation:} magnitude alone conflates a coherent bias with a same-size zero-mean fluctuation, but expectation separates them. Linearizing $F_\theta$ to its Jacobian $J$ ($\lVert J\rVert\le L$), a nonzero mean accumulates to $\E[e_t]\to(I-J)^{-1}\muq(x)$ (magnitude ${\lesssim}\lVert\muq(x)\rVert/(1-L)$), whereas an independent zero-mean perturbation converges to $\E[e_t]\to 0$.
\end{itemize}

This expectation-level separation explains why injecting zero-mean per-step noise fails to rescue Int4 (\Cref{sec:structure:stochastic}) and why per-block scaling succeeds: by shrinking both $\varepsilon$ and $\lVert\muq(x)\rVert$ it restores the contraction regime (MXInt4 $\approx$ FP). The account is conditional. We do not measure $\delta$, so $L+\delta<1$ is inferred from observed drift rather than verified (\Cref{app:theory}). Two controls support it (\Cref{app:dose-untied}): a dose--response sweep where the FP--Int4 gap widens with reuse, and a matched-capacity untied control ($7\times$ less reuse) that roughly doubles Int4 cell-accuracy ($54\%$ vs.\ $25\%$) yet leaves exact accuracy near-zero ($1.6\%$). Weight reuse amplifies the drift, but the root error is the activation-quantization problem of \Cref{sec:blockwise}.

The next section constructs a transition rule that explicitly reduces this bias.

%% file: sections/04_blockwise_integer.tex
\section{Blockwise integer quantization of recursive transitions}
\label{sec:blockwise}

\Cref{sec:timestep} described quantization as a perturbation of a reused transition operator. This section constructs a low-bit transition that stays stable under reuse, guided by a simple principle: control activation-quantization bias locally, at the scale where the recursive transition sees its activation distribution. The resulting rule is a blockwise integer quantizer applied to the activations of the reused transition.

\subsection{Activation quantization dominates transition bias}
\label{sec:blockwise:wa}

We first decompose the 4-bit error into a weight contribution and an activation contribution by quantizing each in isolation (\Cref{fig:ladder} in the appendix). Quantizing weights alone is comparatively benign: on Sudoku it moves the tied model from $84.1\%$ to $78.5\%$ and the untied model from $69.8\%$ to $68.9\%$. The transition changes qualitatively only when activations are quantized as well. Full Int4 (4-bit w$+$a) collapses the tied model to $0.0\%$ and the untied model to $1.6\%$, while 8-bit remains stable throughout (tied Int8 $=84.1\%$). Reuse also compounds the weight-side error: with weight-only Int4 the untied model keeps $99\%$ of FP accuracy versus $93\%$ for the tied model (\Cref{app:dose-untied}). The dominant error term is the activation quantizer inside the reused transition, so a stable low-bit transition must explicitly control activation scales.

\subsection{Granularity controls the transition kernel}
\label{sec:blockwise:granularity}

\begin{figure}[t]
  \centering
  \includegraphics[width=\linewidth]{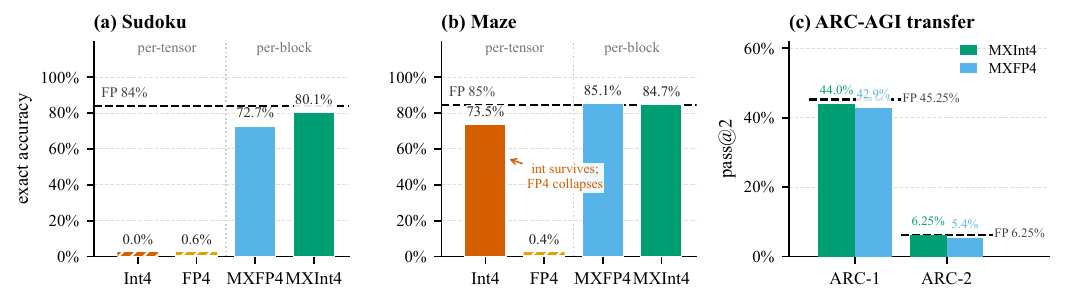}
\caption{\textbf{Granularity, rather than integer vs.\ float, controls stability.} 4-bit format ladder at matched linear-layer coverage (FP embedding). \textbf{(a,b)} On both Sudoku (MLP) and Maze (attention), per-tensor Int4/FP4 collapse (hatched stubs near $0\%$), while per-block MXFP4 (float) and MXInt4 (integer) recover near-FP accuracy---the per-tensor failures span \emph{both} number systems. The Maze exception, per-tensor \emph{integer} surviving ($73.5\%$) while per-tensor \emph{float} collapses ($0.4\%$), is orthogonal to this granularity effect. \textbf{(c)} The same per-block rule transfers to ARC-AGI (pass@2): MXInt4 matches MXFP4 and full precision. MXFP6 and the full ladder (with 3-seed error bars) are in \Cref{tab:transition-quantizers}.}
  \label{fig:formatladder}
\end{figure}

The format ladder separates the number system from scaling granularity. We sweep at fixed coverage, holding the embedding in FP (\Cref{fig:formatladder}). At 4-bit (w$+$a), both per-tensor formats lose accuracy completely (Sudoku Int4 $0.0\%$, FP4 $0.6\%$), while per-block microscaling recovers most of the collapse: MXFP6 returns to full precision ($84.9\% \approx$ FP~$84.1\%$) and MXFP4 recovers substantially to $72.7\%$. Weight-only quantization stays close to full precision across every format, and all 8-bit formats operate at parity. The same split holds on the attention-based Maze task (per-tensor FP4 $0.4\%$, per-block MXFP4/MXFP6 $85.1\%$/$85.4\%$), so the effect extends well beyond Sudoku's MLP recursion. Maze adds one caveat: at \emph{per-tensor} 4-bit the integer format keeps $73.5\%$ while float falls to $0.4\%$, opposite to the usual LLM ordering, where float typically leads at 4-bit. We therefore scope our claim narrowly. The per-tensor failures span integer and (on Sudoku) float formats, per-block scaling brings \emph{both} number systems near full precision, and per-tensor float never exceeds per-tensor integer in our data. Granularity is the mechanism that moves accuracy from near-zero to near-FP, and the residual per-tensor int-vs-float gap on Maze is orthogonal to this primary effect.

Mechanistically, per-block scaling succeeds exactly as \Cref{sec:timestep:mechanism} predicts: giving each block its own scale removes much of the systematic, nonzero-mean transition perturbation, and switching from per-tensor Int4 to per-block MXFP4 flips the Sudoku latent divergence from growing ($270\!\to\!274$) to shrinking ($159\!\to\!129$). Granularity alone controls the contraction-versus-drift boundary of \Cref{sec:timestep:signflip}, which also reconciles our results with the large-LLM MXFP4 consensus (\Cref{app:mxfp4}).

\subsection{Ruling out non-granularity explanations}
\label{sec:blockwise:da1}

If granularity is the operative axis, the alternatives should fail to recover per-tensor Int4, and a sweep over granularity alone should isolate it. We confirm both. \textbf{Calibration freshness.} Replacing the frozen, calibrated per-tensor activation scale with one recomputed online at every recursion step (per-tensor \emph{dynamic} Int4, no calibration) does not recover the collapse: static and dynamic per-tensor Int4 both reach $0.0\%$ on Sudoku, and dynamic also collapses on Maze (to $0.0\%$, below the gracefully degrading static scale of \Cref{sec:timestep:signflip}), whereas per-block MXInt4 recovers full-precision accuracy on both ($80.1\%$/$84.7\%$) with a divergence that \emph{decays} across recursion. A step-varying per-tensor scale re-injects an uncorrelated per-step error that the fixed point cannot absorb. \textbf{Calibration size.} Increasing the \gls{ptq} calibration set $10\times$ ($200\!\to\!2000$ samples) leaves per-tensor Int4 at $0.0\%$ on Sudoku. \textbf{Outlier corrections.} Sweeping SmoothQuant's migration strength and adding a Hadamard rotation leaves it at $\le0.7\%$ (full sweep in \Cref{tab:outliers}). \textbf{Granularity ladder.} Holding 4-bit, dynamic-absmax coverage and per-tensor weights fixed and sweeping only the activation group size, exact accuracy climbs monotonically: $0.0\%$ (per-tensor) $\to 4.6\%$ (per-token) $\to 14.5\%$ (per-group-$32$) $\to 15.0\%$ (per-group-$16$), with nothing else moving, and matched-coverage MXInt4 lifts this ${\sim}15\%$ ceiling to $80.1\%$.\footnote{The ladder holds weights per-tensor and uses a float absmax scale, isolating the \emph{activation}-granularity axis, so its ${\sim}15\%$ ceiling is by design (full MXInt4 scales both weights and activations).} Clipping, freshness, calibration size, and outlier corrections are all ruled out, and spatial granularity is the operative axis.

\Cref{sec:timestep:mechanism} predicted \emph{why}: a coherent bias re-applied at every reuse. We measure this directly by projecting each layer's mean quantization error onto the dominant latent-divergence direction and tracking that direction across recursion steps. The direction is near-constant across steps (cross-step cosine $0.99$ on Sudoku, $0.97$ on Maze) and consistent across examples ($\lVert\overline{d}\rVert/\overline{\lVert d\rVert}\approx0.9$), so the per-step errors accumulate coherently rather than averaging out.

\subsection{The algorithm: blockwise integer transition quantization}
\label{sec:blockwise:mxint}

Per-block float stabilizes the transition, but its per-element floating-point arithmetic is the most expensive part of a microscaling datapath, a real cost in the resource-constrained edge/TinyML regime we target \citep{quanttrim2025,zhuo2022tinymlquant}. We therefore instantiate the same transition-level principle with integer mantissas. For each activation block $B$ of size $32$ we choose a power-of-two shared scale $s_B$, set $q_{\max}=2^{k-1}-1$, and quantize symmetrically to the integer range $[-q_{\max},q_{\max}]$, dequantizing before the next operation:
\begin{equation}
  q_B = \operatorname{clip}\!\left(\Big\lfloor \tfrac{a_B}{s_B}\Big\rceil,\,
  -q_{\max},\, q_{\max}\right), \qquad
  \widehat{a}_B = s_B q_B .
  \label{eq:mxint}
\end{equation}
The scale is local enough to match the activation distribution encountered by the recursive operator, while the mantissa remains integer.

At matched coverage this MXInt quantizer, the \gls{ocp} microscaling \emph{integer} format \citep{rouhani2023mxformats,ocp2023mxspec}, is competitive with per-block float on both task families (\Cref{fig:formatladder}). MXInt4 reaches $80.1\%$ (Sudoku, MLP) and $84.7\%$ (Maze, attention), versus MXFP4's $72.7\%$ and $85.1\%$, both near full precision. Repeating the Sudoku per-block rows across three calibration seeds, MXInt4 is bit-exact at $80.1\%$ (its block scale is the per-block absmax computed online, so run-to-run variance is zero by construction), while MXFP4 is $72.7\%\!\pm\!0.4\%$. The $7.4$-point Sudoku gap is ${\sim}17\times$ the MXFP4 spread, so it is not a single-seed artifact. We still read the two as competitive on these tasks (MXFP4 is marginally ahead on Maze) rather than asserting a universal integer-over-float ordering (\Cref{sec:timestep:theory} discusses the grid-bias interpretation).

The two 4-bit per-block formats have identical memory footprint (block-$32$, one shared E8M0 scale, $4.25$ bits/element) and differ only in the compute datapath: MXInt4 uses integer MACs with power-of-two block scales, whereas MXFP4 needs per-element float decoding, and both require per-block scaled accumulation \citep{rouhani2023mxformats,ocp2023mxspec}. We treat the integer datapath as motivation rather than a measured benefit (\Cref{sec:discussion}).

\Cref{tab:transition-quantizers} collects the \gls{ptq} format ladder at matched linear-layer coverage, with 3-seed error bars on the per-tensor rows: 8-bit operates at parity, per-tensor 4-bit fails in both number systems, and per-block scaling (integer or float) preserves the transition on both task families.

\begin{table}[t]
  \centering
\caption{\textbf{PTQ format ladder} (final-step exact accuracy; matched linear-layer w$+$a coverage, FP embedding). Per-tensor w$+$a rows are mean$\pm$std over $3$ seeds (a seed re-draws only the 200-sample activation-calibration set). Every std is $\le0.4$pp, so the 4-bit collapse (rows pinned near zero) is not a seed artifact and no claim-bearing ordering flips. MXInt4 is deterministic given the checkpoint: its block scale is the per-block absmax computed at inference with no calibration state, so it is bit-exact across seeds. MXFP4 is reported over $3$ Sudoku seeds ($72.7\%\!\pm\!0.4$). Its small spread is a calibration-pass detail of our Brevitas MX path rather than a property of the format, and the $7.4$-point MXInt4--MXFP4 Sudoku gap is ${\sim}17\times$ that spread. On Maze, per-tensor Int4 ($73.5\%$) survives while FP4 ($0.4\%$) collapses. Standard outlier controls do not recover per-tensor Int4 on Sudoku (SmoothQuant $0.7\%$, Hadamard/QuaRot $0.0\%$, single-seed). MXFP6 ($^{\dagger}$) is from the broader-coverage ladder. QAT/LSQ is in the appendix.}
  \label{tab:transition-quantizers}
\footnotesize \setlength{\tabcolsep}{4.5pt} \renewcommand{\arraystretch}{1.0}
  \begin{tabular}{@{}llccc@{}}
    \toprule
    \textbf{Quantizer (PTQ)} & \textbf{Activation scaling} & \textbf{Integer MACs} &
    \textbf{Sudoku (MLP)} & \textbf{Maze (attention)} \\
    \midrule
    Int8                  & per-tensor            & \yes & $84.0\%{\scriptstyle\,\pm.1}$ & $83.6\%{\scriptstyle\,\pm.3}$ \\
    FP8 (e4m3)            & per-tensor            & \no  & $84.1\%{\scriptstyle\,\pm.1}$ & $84.6\%{\scriptstyle\,\pm.2}$ \\
    FP8 (e5m2)            & per-tensor            & \no  & $84.0\%{\scriptstyle\,\pm.1}$ & $84.7\%{\scriptstyle\,\pm.2}$ \\
    \midrule
    Int4                  & per-tensor            & \yes & \lowacc{$0.0\%{\scriptstyle\,\pm.01}$} & $73.5\%{\scriptstyle\,\pm.4}$ \\
    FP4                   & per-tensor            & \no  & \lowacc{$0.6\%{\scriptstyle\,\pm.01}$} & \lowacc{$0.4\%{\scriptstyle\,\pm.3}$} \\
    \ourrow
    \textbf{MXInt4} & \textbf{per-block, integer} & \yes & \textbf{80.1\%} & \textbf{84.7\%} \\
    MXFP4           & per-block, float      & \no  & $72.7\%{\scriptstyle\,\pm.4}$ & 85.1\% \\
    MXFP6           & per-block, float      & \no  & ${\sim}84.9\%^{\dagger}$ & ${\sim}85.4\%^{\dagger}$ \\
    \midrule
    FP (reference)  & n/a & \no & 84.1\% & 84.7\% \\
    \bottomrule
  \end{tabular}
\end{table}

\subsection{Quantization-aware training: a secondary route}
\label{sec:blockwise:lsq}

\gls{qat} alone compensates for much of the per-tensor activation loss: naive Int4 \gls{qat} improves Sudoku from $0.0\%$ to $71.8\%\!\pm\!0.3\%$ and Maze from $73.2\%$ to $82.2\%\!\pm\!0.7\%$, so the residual learned-quantizer gain lies almost entirely in the \emph{weights}. We therefore fix our \gls{raq} method to weight-side \gls{lsq} \citep{esser2020lsq}, which adds $+9.3$ points where naive \gls{qat} struggles (Sudoku, $81.1\%\!\pm\!0.6$) and is tied where it already succeeds (Maze, $81.8\%\!\pm\!1.0\%$). Controls confirm the gain comes from the learned weight step size rather than bias precision, and that the earlier attention instability was specific to activation-\gls{lsq} (\Cref{app:qat}, \Cref{tab:lsq}). Despite these \gls{qat} gains, per-block MXInt4 remains the primary post-training mechanism. Per-iteration normalization and cross-iteration distillation do not help at convergence (\Cref{app:qat}).

\subsection{Transfer to ARC-AGI}
\label{sec:blockwise:arc}

The rule also transfers to ARC-AGI, the open-ended abstraction-and-reasoning benchmark this line of work is known for \citep{jolicoeur2025trm}: few-shot grid-transformation tasks with no fixed output size, scored by \emph{pass@2} (solved if either of two attempts matches the target, \Cref{app:config}). We evaluate ARC-AGI-1 and its harder 2025 successor ARC-AGI-2, the strongest test that our account extends beyond structured Sudoku/Maze environments (full table in \Cref{tab:arc}, \Cref{app:arc}, with the cross-architecture summary in panel (c) of \Cref{fig:formatladder}). Per-tensor Int4-\gls{ptq} gives $0.0\%$ on both datasets, while two distinct routes return to full-precision-level pass@2: per-block scaling at post-training (MXInt4, no retraining, $44.0\%$/$6.25\%\ \approx\ $FP) and naive Int4-\gls{qat} ($45.9\%$/$6.25\%\ \approx\ $FP). The same granularity pattern holds. Weight-\gls{lsq} is neutral here (it helps the MLP recursion but not attention, \Cref{sec:blockwise:lsq}). ARC-AGI-2 is the most quantization-sensitive, with Int8 losing about half its pass@2 score. Because all ARC numbers are single-seed and one ARC-AGI-2 task is worth $6.25\%$, we read them strictly as point estimates rather than a calibrated ordering.

Ultimately, stable low-bit recursive inference requires quantizing the reused transition locally enough that activation-scale mismatch does not become a coherent transition bias. Blockwise \emph{integer} scaling (MXInt4) meets this requirement while keeping integer elements and a power-of-two block scale.

%% file: sections/05_recursion_structure.tex
\section{Recursion structure as transition robustness}
\label{sec:structure}

Previous sections characterized \gls{trm}, a finite-step deterministic transition system whose 4-bit degradation is fundamentally an activation-scaling-granularity problem. We now make a broader architectural claim: \emph{a recursive reasoner's quantizability is shaped by its recursion structure}, with depth and reuse modulating its sensitivity. We test this across an axis of architectures. The deepest point on the axis (\gls{eqr}) is the most sensitive, stochasticity alone does not remove the systematic bias, and per-block activation scaling remains the effective intervention throughout.

\begin{figure}[t]
  \centering
  \includegraphics[width=\linewidth]{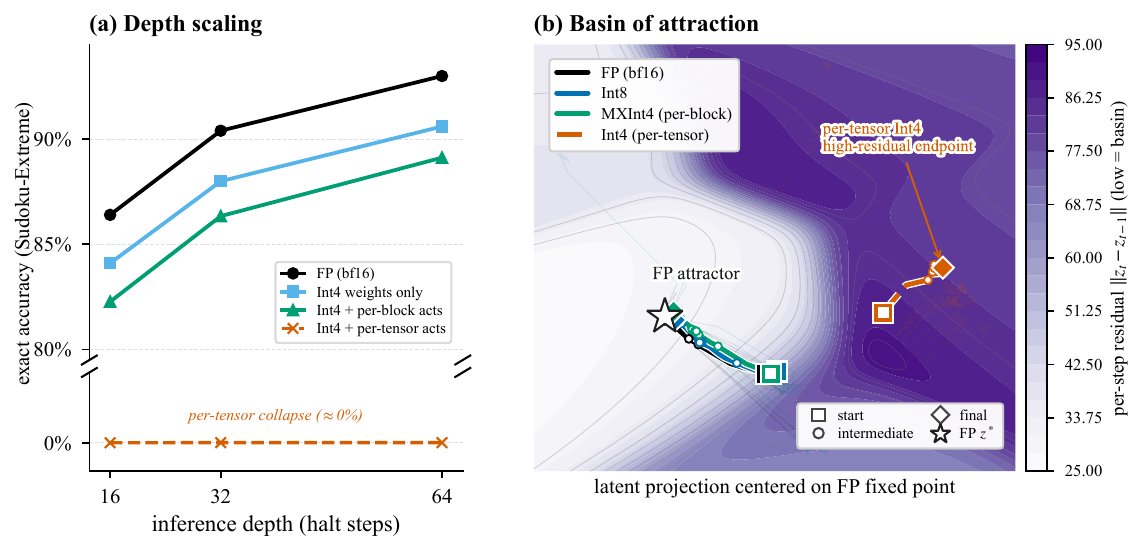}
\caption{\textbf{EqR mechanism: per-block scaling preserves depth use \emph{and} keeps the trajectory in the FP basin.} \textbf{(a)} EqR exact-accuracy vs.\ inference depth: FP and Int4 weight-only scale with depth, per-tensor Int4 \emph{activations} remain near ${\sim}0$, and per-block activations restore both accuracy and depth-scaling ($89.1\%$ at $D64$). \textbf{(b)} FP-centered latent basin: mean trajectory over $12$ Sudoku-Extreme examples projected onto the leading two PCA directions of $z_t-z^*_{\mathrm{FP}}$ (faint lines: per-example paths); the surface is an interpolated per-step residual (low $=$ basin). FP, Int8, and per-block MXInt4 converge into the basin around the FP attractor ($\star$), while per-tensor Int4 terminates in a high-residual region. Squares/circles/diamonds mark first/intermediate/final states. The 2-D projection is illustrative (PC1$+$PC2 ${\approx}30\%$ of variance); contraction vs.\ drift is quantified by the per-step residual and accuracy rather than the projection.}
  \label{fig:eqr-depth}
\end{figure}

\subsection{An axis of recursive transition systems}
\label{sec:structure:axis}

We consider three points on this axis. At the deterministic extreme, \glspl{eqr} iterate a transition to a fixed-point attractor, scaling test-time compute through both \emph{depth} and \emph{breadth} \citep{bai2019deq,huang2026eqr}. In the middle sits the finite-step \gls{trm} studied above. At the stochastic extreme, test-time noise in the style of the \gls{ptrm} or learned per-step transitions (GRAM) explore multiple trajectories \citep{parviz2026ptrm,baek2026gram}. If bit-width were the only factor, all three would behave similarly. Instead, the architecture's required invariant (a fixed point, a finite refinement path, or a distribution over paths) dictates how a systematic bias propagates.

\subsection{Equilibrium recursion is the most sensitive point on the axis}
\label{sec:structure:eqr}

\gls{eqr} pushes recursion depth to the extreme: its released Sudoku-Extreme checkpoint unrolls roughly $288$ layer applications at our base depth (far more at higher depths) and reads its answer from a converged attractor \citep{huang2026eqr}, making it the sharpest test of our accumulation hypothesis: a systematic per-step bias here can displace the entire attractor.\footnote{We use a minimal \texttt{bf16}-consistent fake-quantizer that is FP-neutral (verified FP $86.4\%$). An off-the-shelf harness was not, over ${\sim}288$ applications. \gls{eqr} numbers are for the one released Sudoku-Extreme checkpoint, multi-seed at $D32$ and single-seed at other depths (\Cref{app:config}).}

In full precision \gls{eqr} scales with depth: exact accuracy rises from $86.4\%$ at $D16$ to $93.0\%$ at $D64$ as the final-step residual contracts ($27.5\!\to\!16.2$, \Cref{fig:eqr-depth}). Per-tensor Int4 \emph{activation} quantization destroys this: accuracy drops to effectively zero at every depth ($0.01\%$ at $D16$, $0.02\%\!\pm\!0.002$ across three seeds at $D32$ against FP $90.4\%$) and the equilibrium fails to converge, with the residual stuck near $110$ versus ${\sim}16$ even with a favorable dynamic scale, a massive displacement of the full-precision attractor (\Cref{fig:eqr-depth}b).

As with \gls{trm}, activation-side sensitivity dominates: per-channel weight-only Int4 loses little and keeps depth scaling ($84.1\%\!\to\!90.6\%$ across $D16$--$D64$), and the per-tensor weight-only drop to $5.1\%$ is a coarseness artifact. \gls{eqr}'s stochastic \emph{breadth} also fails to average the activation error away. Aggregation improves full precision (majority vote $91.4\%$, any-correct $93.2\%$) but Int4 stays at ${\sim}0\%$: every trajectory settles into the same shifted basin and agrees on wrong answers, confirming a coherent, systematic bias (\Cref{app:structure}).

Finally, our blockwise rule generalizes to this extreme (\Cref{app:structure}): per-group-$16$ scaling lifts full-Int4 exact accuracy from $0.01\%$ to $82.3\%$ at $D16$ (per-token $60.6\%$, per-group-$32$ $82.2\%$, within ${\sim}4$ points of FP $86.4\%$) and restores depth scaling ($86.3\%$ at $D32$, $89.1\%$ at $D64$). Per-block scaling preserves even the most sensitive model on the axis.

\subsection{Stochasticity does not remove the systematic bias}
\label{sec:structure:stochastic}

At the stochastic end of the axis, test-time stochasticity fails to rescue per-tensor Int4 \gls{trm}: we inject \gls{ptrm}-style Gaussian noise into the deterministic \gls{trm} checkpoint, rather than a trained stochastic model, and sweep $K\in\{1,4,16\}$ trajectories and all noise scales. Both averaging and halt-head selection keep exact accuracy at $0.0\%$, while the same selector lifts full precision to $95.0\%$ (\Cref{app:structure}, \Cref{fig:ptrm}). The same biased transition drives every trajectory, so aggregation cannot recover an unbiased answer. Whether \emph{trained-in} stochasticity (e.g.\ GRAM's learned per-step transitions \citep{baek2026gram}) confers robustness remains open (\Cref{app:structure}).

Across the axis, severity tracks effective recursion depth: the deepest model (\gls{eqr}, ${\sim}288$ applications) is the most sensitive, and no standard defense (stochastic breadth, averaging, halt-head selection, or injected per-step noise) mitigates the drift. Per-block activation scaling is the one intervention that works everywhere.

%% file: sections/07_discussion.tex
\section{Discussion and limitations}
\label{sec:discussion}

\paragraph{A transition-level rule, and what governs it.} We quantize recursive reasoners at the level of their reused transition operator, using local blockwise activation scales with integer mantissas and power-of-two block scales. Per-tensor 4-bit fails because its activation grid induces a biased perturbation re-applied throughout the trajectory, which blockwise scaling reduces locally. More broadly, quantizability depends on more than bit-width and number format. It also depends on the recursive structure that re-applies the transition: how often the operator is reused, whether the answer is read from a finite trajectory or a fixed point, and whether stochastic transitions merely carry variance or absorb systematic bias.

\paragraph{Hardware (motivation only).} MXInt4 and MXFP4 share a $4.25$-bit/element footprint and differ only in the per-element datapath: integer MACs with power-of-two (shift) scales versus per-element float decode. Both still need per-block scaled accumulation, which the integer accelerators that dominate edge silicon (e.g.\ Arm Ethos-U class microNPUs) do not natively provide. Given that support, the integer variant stays all-integer. We frame this as motivation, not a deployment result: a measured kernel and block-aware accumulation support are future work.

\paragraph{Limitations.} Our evaluations cover \gls{trm} on Sudoku-Extreme and Maze-Hard, extended to \gls{eqr}, \gls{ptrm}, \gls{qat}/\gls{lsq}, and ARC-AGI (Sections~\ref{sec:blockwise:lsq}, \ref{sec:blockwise:arc}, and~\ref{sec:structure}). Per-tensor \gls{ptq} and \gls{qat}/\gls{raq} report mean$\pm$std over three seeds, and other measurements are single-seed or deterministic. HRM-scale generality and trained-in stochasticity (GRAM) remain future work. Our microscaling result is regime-specific, consistent with the large-LLM MXFP4 consensus \citep{huawei2026mxfpbenchmark,isfinerbetter2026}. Finally, Proposition~1 is a qualitative scaffold rather than a tight bound or full fixed-point perturbation theory.

%% file: sections/99_appendix.tex
\section{Supporting figures and details}
\label{app:supporting}

This appendix collects supporting evidence referenced in the main text. The main \gls{qat}/\gls{raq} rows (\Cref{tab:lsq}) are reported over three seeds. All remaining tables report single-seed point estimates.

\subsection{Per-step accuracy curves}
\label{app:accuracy}

\begin{figure}[t]
  \centering
  \includegraphics[width=\linewidth]{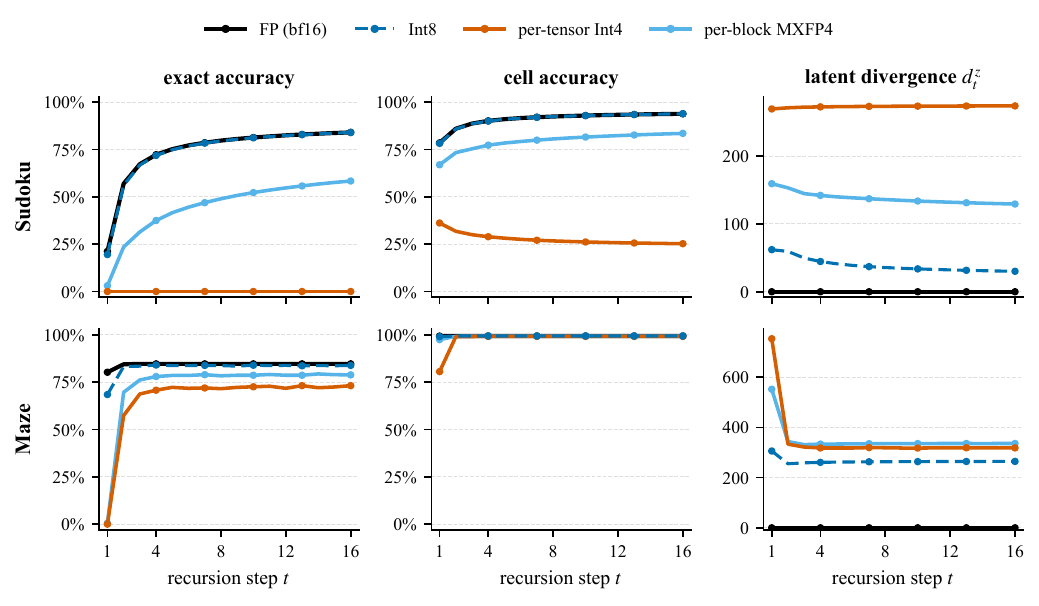}
\caption{\textbf{Per-step trajectory atlas} (full-coverage diagnostic). Exact accuracy, cell accuracy, and latent divergence $d^z_t$ vs.\ recursion step for FP, Int8, per-tensor Int4, and per-block MXFP4, on Sudoku (top) and Maze (bottom). Int8 tracks full precision and its divergence contracts; per-tensor Int4 collapses on the deep Sudoku recursion (flat-high divergence, $0\%$ exact) yet survives the shallow Maze; per-block MXFP4 converts the growing per-tensor divergence into a shrinking one, recovering most accuracy. Int8 is dashed so it does not occlude the FP curve.}
  \label{fig:perstep-atlas}
\end{figure}

\Cref{fig:perstep-atlas} collects the per-step trajectory diagnostic (accuracy and latent divergence) for all four formats, complementing the divergence curves of \Cref{fig:hero}. At 8-bit, accuracy steadily rises toward full precision as the recursion proceeds while the divergence contracts. Conversely, at per-tensor 4-bit, accuracy either fails to improve entirely (Sudoku) or saturates below full precision (Maze).

\subsection{Microscaling stabilizes the activation transition}
\label{app:mxfp4}

\Cref{fig:perstep-atlas} shows the trajectory-level view behind the granularity result of \Cref{sec:blockwise:granularity}: step-by-step, moving from per-tensor Int4 to per-block MXFP4 converts a growing latent divergence into a shrinking one on both tasks.

This connects our result to a current tension in the LLM literature. While 4-bit MXFP is widely reported as lossy and an open challenge \citep{huawei2026mxfpbenchmark,bridginggapmxfp4_2026}, our results show that per-block 4-bit matches or approaches full precision on structured reasoning tasks. The likely reason is the operating regime. Tiny weight-tied models possess narrow, well-behaved weight distributions. These distributions are naturally well-matched to per-block scaling, avoiding the microscaling block-size limits typically reported for large LLMs \citep{isfinerbetter2026}.

\subsection{Per-block integer on Sudoku}
\label{app:mxint-sudoku}

\begin{figure}[h]
  \centering
  \includegraphics[width=0.6\linewidth]{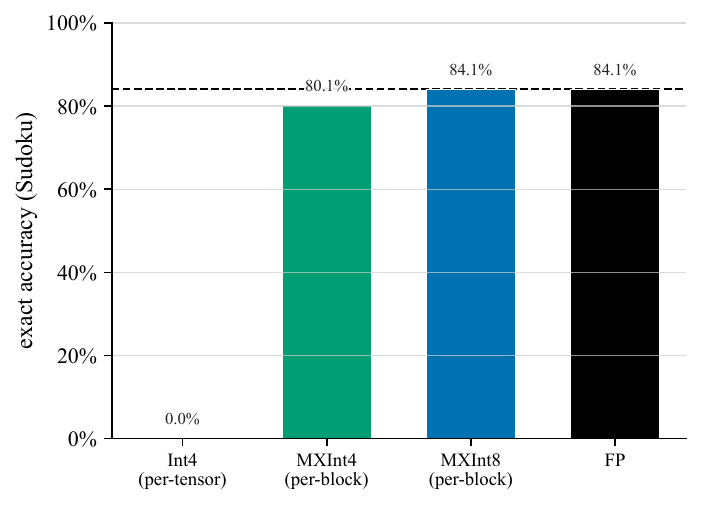}
\caption{Per-block integer (MXInt) results on Sudoku. MXInt8 matches FP; MXInt4 preserves near-FP accuracy where per-tensor Int4 does not.}
  \label{fig:mxint-sudoku}
\end{figure}

\Cref{fig:mxint-sudoku} reports the per-block integer (MXInt) sweep on Sudoku referenced in \Cref{sec:blockwise:mxint}. This includes the MXInt8 gate result ($84.1\%$, matching FP) and the MXInt4 weight-only point ($82.5\%$).

\begin{figure}[t]
  \centering
  \includegraphics[width=0.78\linewidth]{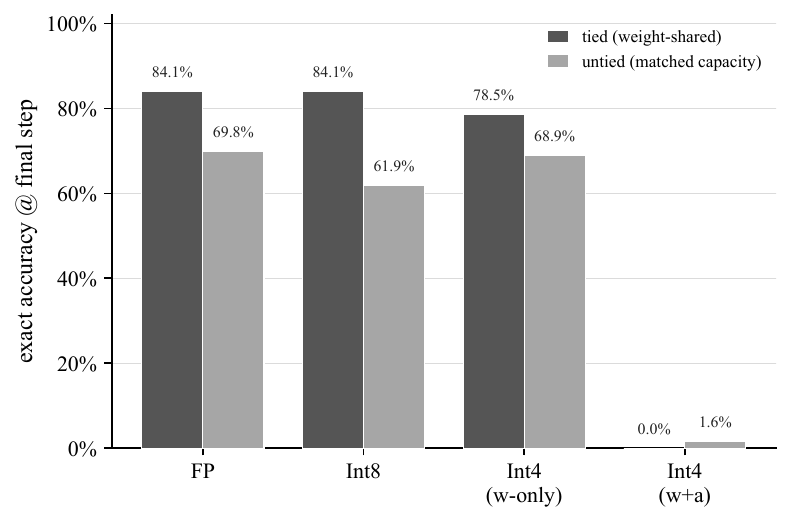}
\caption{\textbf{Activation quantization dominates transition bias.} Weight-vs.-activation decomposition on Sudoku for tied and untied TRM. Weight-only Int4 shifts accuracy only modestly (tied $84.1\%\!\to\!78.5\%$, a ${\sim}7\%$ relative drop); adding activation quantization (4-bit w$+$a) changes the transition enough to remove exact solutions (tied $\to 0.0\%$, untied $\to 1.6\%$). Int8 remains stable.}
  \label{fig:ladder}
\end{figure}

\subsection{Recursion-structure supporting figures}
\label{app:structure}

\begin{figure}[h]
  \centering
  \begin{subfigure}[t]{0.46\linewidth}
    \centering
    \includegraphics[width=\linewidth]{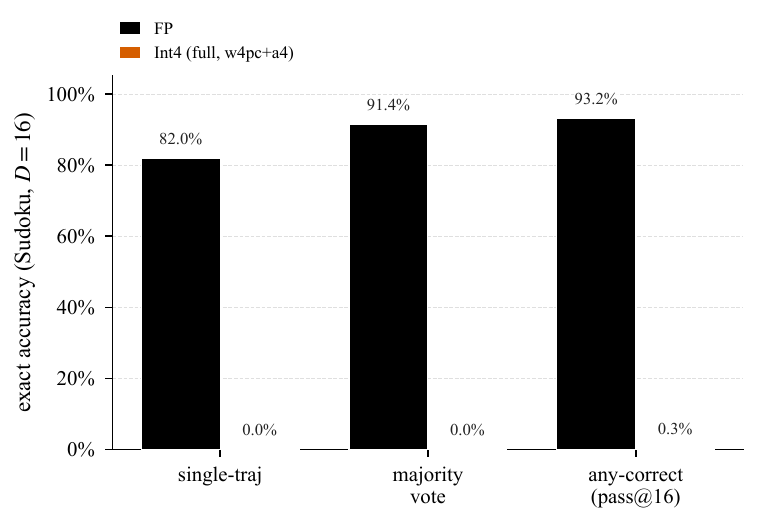}
\caption{EqR: stochastic breadth.}
    \label{fig:eqr-breadth}
  \end{subfigure}
\hfill
  \begin{subfigure}[t]{0.52\linewidth}
    \centering
    \includegraphics[width=\linewidth]{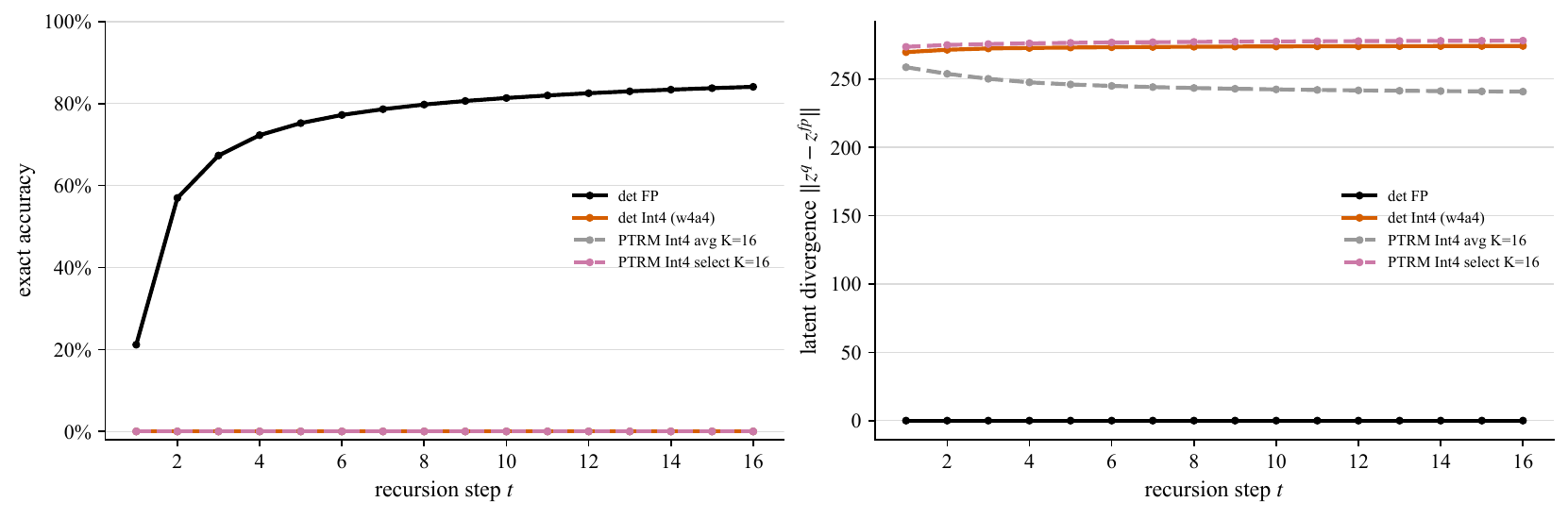}
\caption{PTRM on TRM: per-step accuracy and divergence.}
    \label{fig:ptrm-perstep}
  \end{subfigure}
\caption{Supporting evidence for the recursion-structure analysis. (a) On EqR, breadth improves full precision but not Int4. (b) On TRM, test-time stochasticity leaves Int4 accuracy at zero across all recursion steps.}
  \label{fig:structure-app}
\end{figure}

\Cref{fig:eqr-breadth} provides the \gls{eqr} breadth result behind \Cref{sec:structure:eqr}. Aggregating stochastic trajectories successfully lifts full-precision performance (majority vote $91.4\%$, any-correct $93.2\%$). However, under Int4, accuracy remains pinned at ${\sim}0\%$. Because every trajectory follows the exact same biased transition, averaging simply cannot expose an unbiased path. 

\Cref{fig:ptrm-perstep} provides the per-step view behind \Cref{sec:structure:stochastic}. Under Int4, deterministic and \gls{ptrm} trajectories (whether using averaging or selection) stay at zero accuracy at every single recursion step. Simultaneously, the latent divergence remains trapped near a shifted attractor.

\paragraph{Test-time stochasticity (\gls{ptrm}), in full.} We begin with the near-zero Int4 (4-bit w$+$a) \gls{trm} from \Cref{sec:timestep}. We then inject \gls{ptrm}-style test-time stochasticity \citep{parviz2026ptrm}. This adds zero-mean Gaussian noise at each recursion step, with $K$ trajectories aggregated either by averaging or by selecting the best trajectory using \gls{trm}'s learned halt head. We sweep $K$ and the noise scale without retraining. 

Int4 exact-accuracy stays firmly at $0.0\%$ for every $K\in\{1,4,16\}$ and every noise standard deviation in $\{0.02,\dots,1.0\}$, applied to either latent state. Larger noise only degrades cell-accuracy. Averaging modestly shrinks the latent divergence ($274\!\to\!241$) but fails to improve exact accuracy, while selection simply commits to a single biased trajectory. The control is the exact same machinery on full precision, where halt-head selection lifts accuracy monotonically ($84.1\%\!\to\!89.5\%\!\to\!93.3\%\!\to\!95.0\%$ for $K=1,4,16,64$). The selection mechanism is real and implemented correctly. However, Int4 offers no high-accuracy trajectory to select because every path is corrupted by the same biased transition. 

\paragraph{Trained-in stochasticity (GRAM).} The \gls{ptrm} result specifically addresses stochasticity added \emph{at inference}. Whether \emph{trained-in} stochasticity confers inherent robustness remains an open question for future work. Models like GRAM utilize learned variational per-step transitions \citep{baek2026gram}, but their code and checkpoints are currently unreleased. We are running an in-house proxy to probe this hypothesis (a \gls{trm} trained with stochastic transitions, then quantized). Our working expectation is that training a model to tolerate \emph{zero-mean} variance may not confer tolerance to a \emph{systematic} quantization bias.

\begin{figure}[t]
  \centering
  \includegraphics[width=0.58\linewidth]{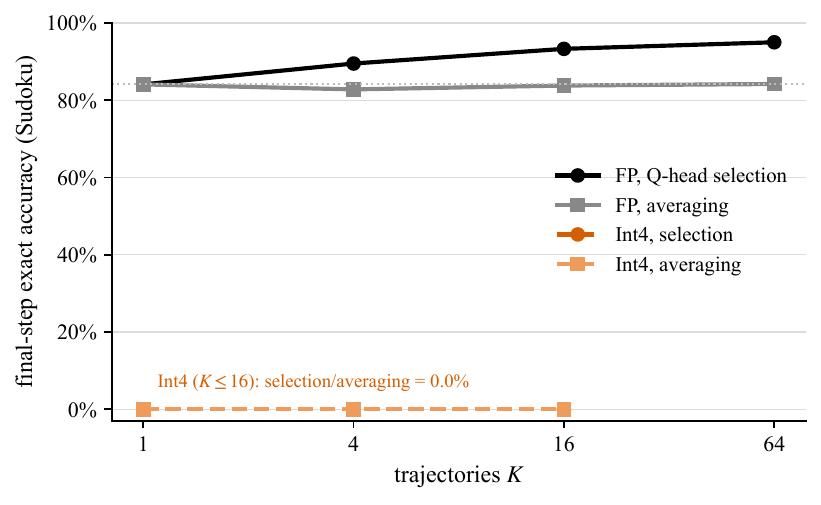}
\caption{\textbf{Test-time stochasticity does not improve Int4 accuracy.} Final-step exact-accuracy vs.\ number of trajectories $K$ on Sudoku (PTRM on TRM, noise $0.05$). On full precision, halt-head selection climbs $84.1\%\!\to\!95.0\%$ (averaging flat at ${\sim}84\%$); on Int4, both averaging and selection remain at $0.0\%$ for all evaluated $K\le16$ (the $K=64$ Int4 run was memory-limited). The perturbation is systematic, so averaging and selection do not expose an unbiased trajectory.}
  \label{fig:ptrm}
\end{figure}

\begin{figure}[t]
  \centering
  \includegraphics[width=0.64\linewidth]{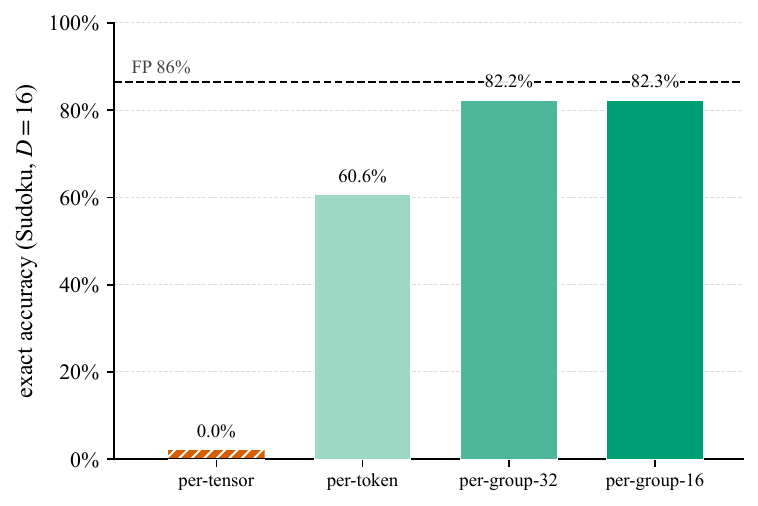}
\caption{\textbf{Activation granularity controls EqR quantization.} Full-Int4 EqR exact-accuracy at $D16$ (weights per-channel) as activation scaling is refined: per-tensor $0.01\%$ $\to$ per-token $60.6\%$ $\to$ per-group-$32$ $82.2\%$ $\to$ per-group-$16$ $82.3\%$, approaching full precision ($86.4\%$). The same per-block principle that preserves TRM and Maze also preserves the extreme-depth equilibrium model.}
  \label{fig:eqr-granularity}
\end{figure}

The basin panel of \Cref{fig:eqr-depth}(b) visualizes the granularity effect as a basin landscape. The surface is a Gaussian-smoothed interpolation of measured per-step residuals over a two-dimensional PCA projection of FP-centered \gls{eqr} states. (This should be read as a trajectory visualization rather than a mathematically fitted potential function). The full-precision path, Int8, and per-block MXInt4 all smoothly move toward the low-residual region around the FP attractor. By contrast, per-tensor Int4 terminates in a high-residual region of the projected state space, despite using the exact same weights as MXInt4. 

\subsection{Reuse controls: dose--response and the untied weight control}
\label{app:dose-untied}

\paragraph{Dose--response in transition reuse.}

\begin{figure}[t]
  \centering
  \includegraphics[width=0.92\linewidth]{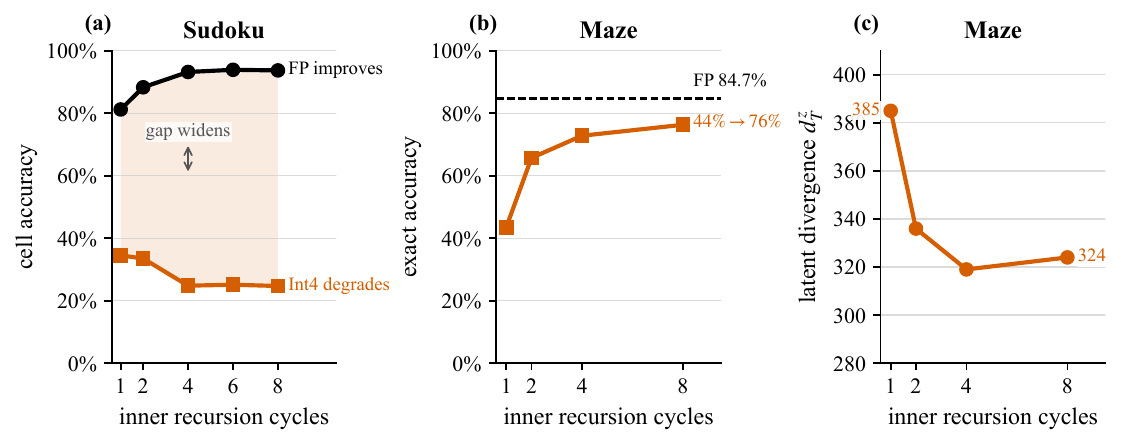}
\caption{\textbf{Dose--response to transition reuse.} Sweeping inner recursion cycles $L$ changes how many times the quantized block is applied before read-out (Sudoku $L\in\{1,2,4,6,8\}$, Maze $L\in\{1,2,4,8\}$). \textbf{(a)} On Sudoku the FP--Int4 cell-accuracy gap \emph{widens} monotonically with reuse (accumulation). \textbf{(b,c)} On the shallower Maze recursion Int4 instead \emph{self-corrects}: exact accuracy climbs toward FP and latent divergence shrinks. Severity tracks effective recursion depth, as the accumulation account predicts.}
  \label{fig:dose}
\end{figure}

If the observed failure is genuinely driven by accumulation, its severity must scale with how often the block is reused. We tested this by sweeping the number of inner recursion cycles, directly multiplying the per-step reuse count. We then measured the full-precision-vs.-Int4 gap at the final step (\Cref{fig:dose}). 

On Sudoku, the gap strictly \emph{widens} with reuse. Full-precision cell-accuracy rises from $81\%$ to $94\%$ as cycles increase. In stark contrast, Int4 cell-accuracy \emph{falls} from $35\%$ to $25\%$, and the Int4 latent divergence jumps once enough cycles are present. Adding recursion makes the full-precision model better and the quantized model worse. This is the signature of error accumulation. On Maze, the same sweep remains safely in the contraction regime: Int4 exact-accuracy climbs from $44\%$ to $76\%$ and latent divergence shrinks ($385\!\to\!324$). The two tasks simply sit on opposite sides of the contraction-versus-drift boundary. The controlling variable dictating success or failure is precisely how many times the quantized block is reused before readout.

\paragraph{Untied matched-capacity control.}

To isolate the effect of weight reuse from the inherent severity of 4-bit quantization, we built a matched-capacity control. This control \emph{unties} the $L$-cycle positions, providing each position with its own distinct weights ($7\times$ less intra-pass reuse). We preserved the rest of the architecture and training pipeline identically. 

\begin{figure}[t]
  \centering
  \includegraphics[width=0.82\linewidth]{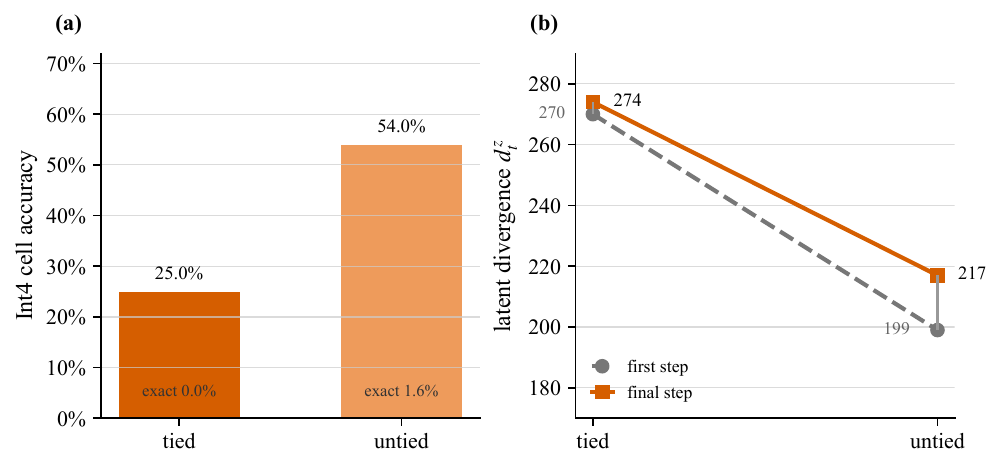}
\caption{\textbf{Matched-capacity untied control.} Untying the $L$-cycle positions ($7\times$ less reuse) roughly doubles Int4 cell-accuracy and lowers final latent divergence. Untied Int4 still has near-zero exact accuracy ($1.6\%$), so reuse amplifies the failure of exact solutions without solely causing it.}
  \label{fig:untied}
\end{figure}

\Cref{fig:untied} provides an informative result in both directions. Reuse clearly \emph{is} a primary driver of the error. Under Int4, the untied model retains roughly twice the cell-accuracy of the tied model ($54\%$ vs.\ $25\%$, or $61\%$ vs.\ $27\%$ of its own full-precision baseline). It also maintains a lower, flatter trajectory divergence ($217$ vs.\ $274$). However, reuse is incomplete as a sole explanation. Even fully untied, exact-accuracy remains near zero ($1.6\%$), and the trajectory divergence still grows over time instead of decaying. 

This happens because the reuse that survives untying is still substantial. Per forward pass, the fully tied model applies each down-projection weight $336$ times. Untying the inner $L$-cycle positions cuts this by $7\times$, resulting in $48$ reuses per weight (contributed by the surviving outer $H$-cycle and deep-supervision loops). Going below $48$ reuses would require completely untying those outer loops as well. That residual reuse of $48$ applications, combined with the intrinsic harshness of per-tensor 4-bit quantization, is enough to destroy exact solutions. Ultimately, weight reuse heavily amplifies transition drift and is necessary to explain the magnitude of the displacement. However, the near-zero accuracy reflects a combination of this reuse and the underlying activation-quantization granularity problem.

\subsection{Quantization-aware training: full results}
\label{app:qat}

Per-block scaling provides a highly effective post-training transition rule. A second, complementary route is \gls{qat}. It is critical to separate this cleanly from the \gls{ptq} results, as the two interact informatively. 

Crucially, \gls{qat} \emph{by itself} compensates for much of the per-tensor activation loss. Naive Int4 \gls{qat} immediately raises Sudoku performance from $0.0\%$ (\gls{ptq}) to $71.8\%\!\pm\!0.3$. It also improves Maze from $73.2\%$ to $82.2\%\!\pm\!0.7$ (\Cref{tab:lsq}). Therefore, within \gls{qat}, much of the activation-scale mismatch is naturally trained away. The residual gain available to a learned quantizer lies entirely in the \emph{weights}. Accordingly, we froze our \gls{qat} method to weight-\gls{lsq}: learning step sizes solely on the weights \citep{esser2020lsq}, while utilizing plain Int4 activations.

This frozen weight-\gls{lsq} (\gls{raq}) method adds ${\sim}9$ points on Sudoku, jumping from $71.8\%\!\pm\!0.3$ to $81.1\%\!\pm\!0.6$. This gap far exceeds per-seed spread and closes the majority of the remaining FP gap. A targeted decomposition confirms this gain lives strictly in the weights. Applying \gls{lsq} solely to activations yields $70.9\%$ (essentially the naive baseline). Conversely, applying it to the weights yields the full $81.1\%$. Furthermore, an ablation that varies only the bias precision leaves accuracy completely unchanged, showing the gain stems from the learned step size rather than the bias representation. 

This picture is remarkably consistent across architectures. On Maze's attention recursion, \gls{raq} is statistically tied with naive \gls{qat} and highly stable ($81.8\%\!\pm\!1.0$ vs.\ $82.2\%\!\pm\!0.7$, both near the FP $84.7\%$ baseline). The instability observed earlier was specific to quantizing the attention \emph{activations} (activation-\gls{lsq} drops performance to $64.9\%$). The frozen weight-side method explicitly avoids this trap. 

In summary, weight-\gls{lsq} delivers a significant win where naive \gls{qat} struggles (Sudoku/MLP), while remaining stable where naive \gls{qat} already succeeds (Maze/attention). Despite these \gls{qat} gains, per-block scaling (\Cref{tab:transition-quantizers}) remains the superior and primary post-training mechanism. 

\begin{table}[t]
  \centering
\caption{\textbf{Quantization-aware training: a secondary, weight-side route} ($12$k steps). QAT alone already compensates for much of the per-tensor activation loss (naive Int4 QAT, Sudoku $0.0\%\!\to\!71.8\%$); the residual learned-quantizer gain is in the \emph{weights} (weight-LSQ $=$ our RAQ): a significant $+9.3$ points on Sudoku (gap $\gg$ std) and statistically tied on Maze, a cross-architecture result. The naive-QAT and RAQ rows are mean$\pm$std over $3$ seeds; the decomposition rows are single-seed. Per-block MXInt4 remains the primary post-training mechanism.}
  \label{tab:lsq}
\footnotesize \setlength{\tabcolsep}{9pt} \renewcommand{\arraystretch}{1.15}
  \begin{tabular}{@{}lcc@{}}
    \toprule
    \textbf{Method (4-bit)} & \textbf{Sudoku (MLP)} & \textbf{Maze (attention)} \\
    \midrule
    Int4 PTQ (eval-only)         & \lowacc{0.0\%} & 73.2\% \\
    naive Int4 QAT               & $71.8\%$\std{0.3} & $82.2\%$\std{0.7} \\
    \ourrow
    \textbf{$+$ weight-LSQ (RAQ)} & \textbf{$81.1\%$}\std{0.6} & $81.8\%$\std{1.0} \\
    \quad activation-LSQ only    & 70.9\% & \lowacc{64.9\%} \\
    \midrule
    FP (reference)               & 84.1\% & 84.7\% \\
    \bottomrule
  \end{tabular}
\end{table}

\paragraph{Per-iteration alternatives (E4/E5): negative results.}

The transition-bias theory suggests two natural per-iteration defenses. The first is iteration-decoupled normalization (E4), which gives the reused block per-iteration normalization statistics so the quantized activation distribution remains stationary. The second is cross-iteration distillation (E5), which distills the quantized recursion toward the full-precision trajectory step-by-step \citep{yao2022zeroquant,zhu2023qfd}. Both naturally target the per-iteration bias. 

However, at convergence, neither method improves upon the baseline. On Sudoku, E4 reaches $70.7\%$, E5 reaches $69.4\%$, and their combination reaches $72.7\%$. All three are numerically identical to the matching single-seed naive-\gls{qat} baseline ($72.2\%$; the three-seed mean is $71.8\%\!\pm\!0.3$, \Cref{tab:lsq}). We therefore report these strictly as negative results. Once the activation scale is handled properly, the per-iteration structure requires no special algorithmic handling. Per-block scaling remains the simplest, most robust transition-level rule.

\subsection{Static outlier controls do not rescue per-tensor Int4}
\label{app:outliers}

\Cref{tab:outliers} reports the full outlier-control sweep behind \Cref{sec:blockwise:da1}. A natural objection is that per-tensor Int4 fails for the usual reason transformer activations are hard to quantize: a few large outlier channels. We test two canonical activation-outlier methods at matched coverage on Sudoku (the maximum-headroom case): SmoothQuant, which migrates activation scale into the weights \citep{xiao2023smoothquant}, swept across its migration strength $\alpha$, and a fixed Hadamard rotation as in QuaRot \citep{ashkboos2024quarot}. Neither closes the gap: the best SmoothQuant setting ($\alpha{=}0.5$) reaches only $0.7\%$ and Hadamard $0.0\%$, versus MXInt4's $80.1\%$. This matches our accumulation account. A \emph{static} per-channel scale or \emph{fixed} rotation corrects the outlier distribution only once; after that the divergence stays high and flat across steps (SmoothQuant $242\!\to\!234$, Hadamard $258\!\to\!245$) instead of contracting. To succeed, the activation scale must dynamically track local block ranges at each recursive application.

\begin{table}[t]
  \centering
\caption{\textbf{Static activation-outlier controls do not rescue per-tensor Int4} (Sudoku, $4$-bit w$+$a, qlinear-only coverage, final step; single-seed). SmoothQuant is swept over its migration strength $\alpha$; QuaRot applies a fixed Hadamard rotation. The best static correction ($\alpha{=}0.5$) reaches $0.7\%$, versus $80.1\%$ for per-block MXInt4.}
  \label{tab:outliers}
\footnotesize \setlength{\tabcolsep}{10pt} \renewcommand{\arraystretch}{1.15}
  \begin{tabular}{@{}lc@{}}
    \toprule
    \textbf{Method (Sudoku, w4a4)} & \textbf{Exact accuracy} \\
    \midrule
    per-tensor Int4 (no correction)   & \lowacc{$0.0\%$} \\
    SmoothQuant $\alpha{=}0.3$         & \lowacc{$0.0\%$} \\
    SmoothQuant $\alpha{=}0.5$ (best)  & \lowacc{$0.7\%$} \\
    SmoothQuant $\alpha{=}0.7$         & \lowacc{$0.4\%$} \\
    SmoothQuant $\alpha{=}0.8$         & \lowacc{$0.03\%$} \\
    SmoothQuant $\alpha{=}0.85$        & \lowacc{$0.0\%$} \\
    SmoothQuant $\alpha{=}0.95$        & \lowacc{$0.0\%$} \\
    Hadamard rotation (QuaRot)        & \lowacc{$0.0\%$} \\
    \midrule
    per-block MXInt4 (reference)      & \textbf{$80.1\%$} \\
    \bottomrule
  \end{tabular}
\end{table}

\subsection{Iterated models and transition-aware quantization}
\label{app:diffusion}

The mechanism we study, per-step error compounding across an iterated computation, is well documented in other iterated domains. In autoregressive and diffusion generation, exposure-bias research frames train--test mismatch as per-step errors accumulating across a generated trajectory \citep{ning2023exposurebias,arora2022exposurebias}. Self-recovery analyses similarly caution that autoregressive distortions need not grow monotonically \citep{he2021quantifyingexposure}. A vast body of mitigation work addresses this specific mismatch through input perturbation, scheduled sampling, or shifted timesteps \citep{ning2023inputperturbation,bengio2015scheduledsampling,li2024tsdpm}.

\subsection{Diffusion quantization and manifold drift}
\label{sec:related:diffusion}

Quantized diffusion models provide the closest theoretical analogue to our timestep-accumulation effect. A diffusion sampler repeatedly queries a single denoising network with shared weights, and uses the result to sequentially update a latent trajectory. Consequently, \gls{ptq} perturbs an iterated transition kernel rather than a collection of independent feedforward layers. 

PTQ4DM and Q-Diffusion successfully identify this multi-timestep structure as the central obstacle. Because activation and noise-prediction distributions shift with the timestep, a single-step calibration unavoidably induces step-dependent bias \citep{shang2023ptq4dm,li2023qdiffusion}. Temporal Dynamic Quantization and TFMQ-DM make this exact dependence explicit through timestep-conditioned statistics \citep{so2023tdq,huang2024tfmqdm}. PTQD provides a full dynamical account: quantization noise decomposes into correlated and residual components, induces mean deviation, and accumulates over denoising steps to drastically reduce the signal-to-noise ratio (SNR) at late timesteps \citep{he2023ptqd}. QNCD and TAC further distinguish single-step disturbance from inter-step accumulation, framing the latter as exposure bias across the sampling trajectory \citep{chu2024qncd,yao2024tacdiffusion}.

This represents the continuous generative counterpart of our 4-bit trajectory drift. In a diffusion sampler, systematic quantization shifts the estimated mean update. As a result, the reverse trajectory progressively abandons the full-precision path and drifts away from the data manifold. Recent methods combat this by treating quantization as a backward displacement \citep{chen2024stepbaq}, explicitly altering the sampling trajectory \citep{frumkin2025qsched}, or applying a marginal-preserving drift adjustment \citep{ryu2026qdrift}. 

In a recursive reasoner, the mathematical equivalent is a refinement path through latent logic space. Our per-tensor Int4 result is the direct logical analogue of this manifold drift. The quantized transition repeatedly injects a biased activation perturbation, pushing the reasoning trajectory away from the full-precision solution manifold instead of allowing it to contract back.

The SNR comparison highlights exactly why this effect is so catastrophic for reasoning models. Diffusion samplers are most fragile late in the reverse chain, when the residual update is small and quantization noise easily dominates microscopic refinements. Recursive reasoners possess an analogous late-stage regime. Their terminal states are strictly constrained by discrete global validity: a Sudoku assignment must perfectly obey all row, column, and box constraints. In continuous image space, modest manifold drift merely manifests as blur or local artifacts. In discrete logical space, the threshold is sharp. A small, systematic displacement changes a single symbol, violating a global constraint and instantly reducing exact accuracy to zero. Both systems accumulate timestep error under iterated shared weights. However, while the perceptual manifold tolerates small geometric deviations, the logical solution manifold is effectively all-or-nothing.

\subsection{Mechanism: per-layer quantization bias}
\label{app:mechanism}

\begin{figure}[t]
  \centering
  \includegraphics[width=0.78\linewidth]{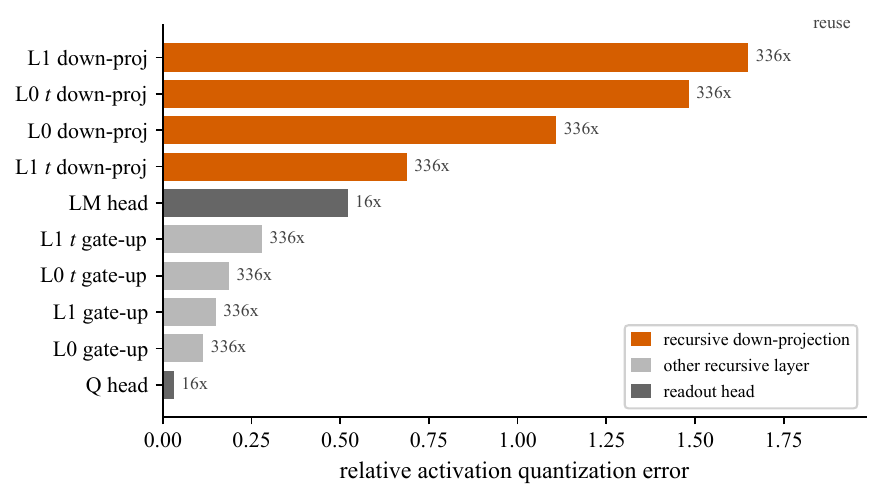}
\caption{\textbf{Reused transition layers dominate activation error.} Per-layer relative activation error magnitude for a calibrated Int4 TRM on Sudoku. The largest errors occur in recursive down-projection layers that are re-applied $336\times$ per forward pass, compared with $16\times$ for the readout heads. This localizes the transition components where any systematic quantization component is repeatedly injected.}
  \label{fig:mechanism}
\end{figure}

\Cref{fig:mechanism} details the per-layer calibration data supporting \Cref{sec:timestep:mechanism}. The relative activation quantization error concentrates almost entirely in the recursive down-projection layers. These are exactly the components the recursion re-applies most frequently per forward pass. This directly localizes the transition components responsible for repeatedly injecting the systematic quantization bias.

\subsection{Dynamical-systems perspective: full statement}
\label{app:theory}

The contraction-versus-drift phenomenon admits a lightweight, conditional mathematical characterization. To formulate this, we must cleanly separate two properties of the per-step transition perturbation $\epsilon_Q$ from \Cref{eq:q-transition}: its \emph{magnitude} and its \emph{state-variation}. On a closed neighbourhood $U$ containing the trajectory region visited by the recursion, and for a fixed input $x$, we define:
\begin{equation}
  \sup_{s\in U}\lVert\epsilon_Q(s,x)\rVert \le \varepsilon, \qquad
  \lVert\epsilon_Q(s,x)-\epsilon_Q(s',x)\rVert \le \delta\,\lVert s-s'\rVert, \qquad
  \muq(x) \coloneqq \E_{t}\!\big[\epsilon_Q(s_t^q,x)\big]
  \label{eq:decomp}
\end{equation}
Here, $\varepsilon$ strictly bounds the magnitude and $\delta$ bounds the state-variation over $U$. Crucially, $\muq(x)$ isolates the systematic, nonzero-mean component of the perturbation \emph{along input $x$'s specific trajectory}. (This is an average over steps, which uniquely drives the per-input displacement. It is not an input-pooled average, which could falsely cancel out coherent biases pointing in different directions for different inputs). 

Because a real quantizer is piecewise-constant, and therefore not literally Lipschitz continuous at its rounding boundaries, we read $\delta$ as a \emph{smoothed} local sensitivity. It acts as an expected, non-pointwise variation bound on $\epsilon_Q$. \Cref{sec:timestep:mechanism} measures the empirical layerwise analogue of $\muq$. 

Empirically, the full-precision recursion is highly contractive near a fixed point $s^\ast$. Its residuals decay dramatically over time (e.g., on \gls{eqr}, the FP residual falls $27.5\!\to\!16.2$ with depth; at 8-bit, the divergence in \Cref{fig:hero} rapidly shrinks). Consequently, we \emph{model} $F_\theta$ as a contraction with modulus $L<1$ near $s^\ast$. This is supported by the measured decay, though we do not claim it as formally proven.

\begin{proposition}[Local perturbed fixed-point model; conditional]
\label{prop:accum}
Fix an input $x$, and let $F_\theta(\cdot,x)$ be locally contractive with modulus $L<1$ on a closed ball $U$ around a fixed point $s^\ast$. Let $\widetilde F = F_\theta + \epsilon_Q$, with $\epsilon_Q$ satisfying the bounds in \Cref{eq:decomp} on $U$. If $L+\delta<1$ and $\widetilde F$ maps $U$ into itself, then $\widetilde F$ has a unique fixed point $\widetilde s^\ast\in U$ (by Banach's fixed-point theorem).\footnote{Contraction: $\lVert\widetilde F(s,x)-\widetilde F(s',x)\rVert\le(L+\delta)\lVert s-s'\rVert$, so $L+\delta<1$ gives a unique fixed point by Banach's theorem. Displacement: with $s^\ast=F_\theta(s^\ast,x)$ and $\widetilde s^\ast=F_\theta(\widetilde s^\ast,x)+\epsilon_Q(\widetilde s^\ast,x)$, contraction gives $\lVert\widetilde s^\ast-s^\ast\rVert\le L\lVert\widetilde s^\ast-s^\ast\rVert+\varepsilon$, hence \Cref{eq:displacement}. In the equilibrium limit this matches the displacement-versus-margin bound of \citet{moneq2026}.} Its iterates converge to this point, and:
\begin{equation}
  \lVert \widetilde s^\ast - s^\ast\rVert \le \frac{\varepsilon}{1-L}.
  \label{eq:displacement}
\end{equation}
\end{proposition}

\noindent\emph{Remark (informal).} Under the further idealization that $F_\theta$ is linearized to a fixed Jacobian $J$ along the trajectory ($\lVert J\rVert\le L$) and the per-step perturbation carries a constant mean $\muq(x)$, the trajectory error acquires a coherent component of order $\lVert\muq(x)\rVert/(1-L)$. Fluctuations of comparable size that are zero-mean and \emph{independent across steps} act purely through their variance and produce no deterministic displacement in expectation.

These two quantities govern two entirely distinct failure modes. The \emph{magnitude} $\varepsilon$ explicitly sets the fixed-point displacement $\varepsilon/(1-L)$. The \emph{state-variation} $\delta$ (via $L+\delta<1$) dictates whether the system remains contractive at all. 

When both terms are small, the model enters the 8-bit regime: it suffers a minor displacement but degrades gracefully (Int8 $\approx$ FP). Per-tensor Int4, however, represents the high-error regime. The observed divergence growth signifies a perturbation that either violently shifts the local attractor, pushes the trajectory entirely outside the locally contractive region, or outright destroys local contraction. This is consistent with the \gls{eqr} residual sticking at ${\sim}110$ (versus FP ${\sim}16$) and the exploding 4-bit divergence seen in \Cref{fig:hero}. Under this joint-contraction model, both components of $s=(z,y)$ should eventually saturate. Empirically, the latent divergence $d^z_t$ does saturate near $270$. The answer-logit divergence $d^y_t$ continues climbing through step~$16$. This suggests the readout converges more slowly, hasn't reached its own perturbed fixed point, and acts as an amplifier for the persistent latent bias. 

\paragraph{Why a systematic bias is the culprit.} Let $e_t = s_t^q - s_t^{\mathrm{fp}}$ represent the trajectory error. The local $L$-contraction of $F_\theta$ provides the strict magnitude bound $\lVert e_{t+1}\rVert \le L\lVert e_t\rVert + \lVert\epsilon_Q(s_t^q,x)\rVert$. Unrolling this with $\lVert\epsilon_Q\rVert\le\varepsilon$ yields:
\begin{equation}
  \lVert e_t\rVert \;\lesssim\; L^t\lVert e_0\rVert + \frac{1-L^t}{1-L}\,\varepsilon
  \;\xrightarrow[t\to\infty]{}\; \frac{\varepsilon}{1-L}.
  \label{eq:accum}
\end{equation}
This bound fundamentally conflates a coherent bias with a zero-mean fluctuation. Both simply contribute $\varepsilon$ to the right-hand side. The true difference materializes entirely in \emph{expectation}. If we linearize $F_\theta$ along the trajectory to its Jacobian $J$ ($\lVert J\rVert\le L$), a coherent mean $\muq(x)$ forces $\E[e_t]\to (I-J)^{-1}\muq(x)$. This carries a magnitude $\lesssim\lVert\muq(x)\rVert/(1-L)$. By contrast, a perturbation that is zero-mean and independent across steps yields $\E[e_t]\to 0$. Only its variance accumulates to a bounded steady state. 

Because the recursive reasoner repeatedly applies the exact same quantized transition with the exact same parameters, a nonzero $\muq(x)$ is continuously injected in a consistent direction. This generates a deterministic displacement. Zero-mean fluctuations do not. This separation in expectation explains why injecting zero-mean noise at test time fails to rescue Int4 accuracy (\Cref{sec:structure:stochastic}). It also explains why per-block scaling is the correct intervention: it shrinks both the magnitude $\varepsilon$ and the systematic bias $\lVert\muq(x)\rVert$, pushing the dynamics back into the contraction regime. 

\paragraph{Scope.} \Cref{prop:accum} remains a \emph{conditional local model}. We do not assert that \gls{trm} or \gls{eqr} are globally contractive. The contraction assumption is derived directly from empirical residual decay. Furthermore, two specific idealizations are present. First, the Lipschitz bound cannot hold pointwise for a quantizer, so $\delta$ must be treated as an expected (smoothed) sensitivity. In particular, the literal piecewise-constant map need not admit an exact fixed point, and its iterates may instead settle into a small cycle. The empirical plateaus indicate convergence into a small neighborhood, which is what the smoothed model describes. Second, we do not directly estimate $\delta$. We measure the empirical analogues of $\varepsilon$ (relative error), $L$ (residual decay), and $\muq$ (layerwise mean), but the regime split $L+\delta<1$ is inferred through observed behavioral drift rather than checked mathematically. We therefore present this model as strong qualitative corroboration rather than a strict quantitative fit. Formulating a full, non-idealized nonlinear perturbation theory for discrete reasoners remains exciting future work.

\subsection{ARC-AGI transfer: full table}
\label{app:arc}

Full pass@2 results supporting \Cref{sec:blockwise:arc}. The cross-architecture summary is also visualized in panel (c) of \Cref{fig:formatladder}.

\begin{table}[t]
  \centering
\caption{\textbf{Both quantization routes transfer to ARC-AGI} (pass@2). Per-tensor Int4-PTQ gives $0\%$ on both datasets, while two independent routes reach the FP-level point estimate: per-block scaling at post-training (MXInt4, no retraining) and quantization-aware training (naive Int4-QAT). On ARC-AGI-2, 4-bit MXInt4 matches the FP point estimate and lands above 8-bit per-tensor Int8 (single-seed, and one ARC-AGI-2 task is worth $6.25\%$). MX entries use matched linear-layer coverage.\protect\footnotemark}
  \label{tab:arc}
\footnotesize \setlength{\tabcolsep}{10pt} \renewcommand{\arraystretch}{1.2}
  \begin{tabular}{@{}llcc@{}}
    \toprule
    \textbf{Quantizer} & \textbf{Scaling} &
    \textbf{ARC-AGI-1} & \textbf{ARC-AGI-2} \\
    \midrule
    FP (reference)        & n/a                      & 45.25\% & 6.25\% \\
    Int8                  & per-tensor, 8-bit (PTQ)  & 41.6\%  & 2.9\%  \\
    Int4                  & per-tensor, 4-bit (PTQ)  & \lowacc{0.0\%} & \lowacc{0.0\%} \\
    MXFP4                 & per-block float (PTQ)    & 42.9\% & 5.4\% \\
    \ourrow
    \textbf{MXInt4}       & \textbf{per-block integer (PTQ)} & \textbf{44.0\%} & \textbf{6.25\%} \\
    \midrule
    naive Int4-QAT        & per-tensor (QAT)         & 45.9\% & 6.25\% \\
    $+$ weight-LSQ (RAQ)  & learned weights (QAT)    & 44.6\% & 4.86\% \\
    \bottomrule
  \end{tabular}
\end{table}
\footnotetext{Full-coverage MXFP4 is omitted because it is not coverage-matched to MXInt4 and so would not be an apples-to-apples comparison; both MX entries here use identical matched linear-layer weight+activation coverage.}

\subsection{Experimental configuration}
\label{app:config}

\paragraph{Datasets and metrics.} \emph{Sudoku-Extreme} \citep{wang2025hrm,jolicoeur2025trm} comprises hard $9\times9$ Sudoku puzzles encoded as length-$81$ token sequences (one token per cell); we evaluate on the full $422{,}786$-puzzle test split (models are trained with $1000$ augmentations per puzzle). \emph{Maze-Hard} comprises $30\times30$ grid mazes (length-$900$ sequences) that require a valid path from start to goal. For both, \emph{exact accuracy} is the fraction of test instances solved \emph{completely} correctly (all $81$ cells, or the entire path), whereas \emph{cell accuracy} is the fraction of individual cells/positions correct, a finer-grained signal that can stay high even when no full solution is produced (e.g.\ a near-miss Sudoku grid). \emph{ARC-AGI-1} and its harder 2025 successor \emph{ARC-AGI-2} are open-ended abstraction-and-reasoning benchmarks of few-shot grid-transformation tasks with no fixed output size; following the recursive-reasoning line of work \citep{jolicoeur2025trm} we report \emph{pass@2} (a task counts as solved if either of two attempts, produced by test-time-augmentation voting, matches the target). Our ARC-AGI-2 evaluation subset is small enough that a single task is worth $6.25\%$, so those numbers are read as point estimates rather than a calibrated ordering.

Unless stated otherwise, results utilize the public Sudoku-Extreme and Maze-Hard benchmarks alongside \gls{trm} models trained using the publicly released codebase. (The tied FP baseline is $84.1\%$ on Sudoku and $84.7\%$ on Maze; the untied control FP baseline is $69.8\%$ on Sudoku). 

\gls{eqr} evaluations use the released Equilibrium Reasoner Sudoku-Extreme checkpoint \citep{huang2026eqr} paired with a \texttt{bf16}-consistent fake-quantizer. (This quantizer is FP-neutral, verified at an accuracy of $86.4\%$). The recursion uses \texttt{halt\_max\_steps}~$=16$. The dose--response sweep specifically varies the inner $L$-cycles across $\{1,2,4,6,8\}$. 

\paragraph{Quantizer coverage across protocols.} Per-step trajectory diagnostics (\Cref{fig:hero} and the dynamic/granularity controls of \Cref{sec:blockwise:da1}) quantize the linear layers of the recursive block and keep all other modules in full precision (single seed). The format ladder of \Cref{tab:transition-quantizers} uses a slightly broader integer coverage with three calibration seeds. Endpoint accuracies therefore differ slightly between the two protocols (e.g.\ Sudoku Int8 $84.2\%$ in \Cref{fig:hero} vs.\ $84.0\%\!\pm\!0.1$ in the ladder; Maze per-tensor Int4 $77.7\%$ vs.\ $73.5\%\!\pm\!0.4$). Within any one figure or table, coverage is always matched across the formats being compared.

Quantization is applied via the configuration interface outlined in the released code. To ensure strict apples-to-apples comparisons, the embedding is held in full precision across all evaluated formats. Per-block formats utilize a block size of $32$ combined with a power-of-two (E8M0) shared scale. Full hyperparameter configurations, exact random seeds, and comprehensive per-step logging scripts are provided in the supplementary material.